\documentclass[pdflatex,sn-mathphys-num]{sn-jnl}% Math and Physical Sciences Numbered Reference Style
\usepackage{longtable}    % Для многостраничных таблиц
\usepackage{tabularx}     % Для контроля ширины колонок
\usepackage{caption}      % Для улучшения подписей
\usepackage{array}        % Для задания фиксированных размеров ячеек
\usepackage{graphicx}
\usepackage{multirow}
\usepackage{amsmath,amssymb,amsfonts}
\usepackage{amsthm}
\usepackage{mathrsfs}
\usepackage[title]{appendix}%
\usepackage{xcolor}%
\usepackage{textcomp}%
\usepackage{manyfoot}%
\usepackage{booktabs}%
\usepackage{algorithm}%
\usepackage{subcaption}
\usepackage{algorithmicx}%
\usepackage{algpseudocode}%
\usepackage{listings}%
\usepackage[dvipsnames]{xcolor}
\graphicspath{{./images/}}
\theoremstyle{thmstyleone}%
\theoremstyle{thmstyletwo}%

\theoremstyle{thmstylethree}%

\begin{document}

\title[Article Title]{Fuzzy Expert System for the Process of Collecting and Purifying Acidic Water: A Digital Twin Approach}


\author*[1]{\fnm{Temirbolat} \sur{Maratuly}}\email{t.maratuly@kbtu.kz}
\author*[1]{\fnm{Pakizar} \sur{Shamoi}}\email{p.shamoi@kbtu.kz}
\author[1]{\fnm{Timur} \sur{Samigulin}}\email{t.samigulin@kbtu.kz}

% \author[1]{\fnm{Third} \sur{Author}}\email{iiiauthor@gmail.com}

\affil*[1]{\orgdiv{School of Information Technology and Engineering}, \orgname{Kazakh-British Technical University}, \orgaddress{\street{Tole Bi Street, 59}, \city{Almaty}, \postcode{050000}, \country{Kazakhstan}}}

% \affil[2]{\orgdiv{Department}, \orgname{Organization}, \orgaddress{\street{Street}, \city{City}, \postcode{10587}, \state{State}, \country{Country}}}

% \affil[3]{\orgdiv{Department}, \orgname{Organization}, \orgaddress{\street{Street}, \city{City}, \postcode{610101}, \state{State}, \country{Country}}}


\abstract{Purifying sour water is essential for reducing emissions, minimizing corrosion risks, enabling the reuse of treated water in industrial or domestic applications, and ultimately lowering operational costs. Moreover, automating the purification process helps reduce the risk of worker harm by limiting human involvement. Crude oil contains acidic components such as hydrogen sulfide ($\text{H}_{2}\text{S}$), carbon dioxide ($\text{CO}_{2}$), and other chemical compounds. During processing, these substances are partially released into sour water. If not properly treated, sour water poses serious environmental threats and accelerates the corrosion of pipelines and equipment.  This paper presents a fuzzy expert system, combined with a custom-generated digital twin, developed from a documented industrial process to maintain key parameters at desired levels by mimicking human reasoning. The control strategy is designed to be simple and intuitive, allowing junior or non-expert personnel to interact with the system effectively. The digital twin was developed using Honeywell’s UniSim Design R492 to simulate real industrial behavior accurately. Valve dynamics were modeled through system identification in MATLAB, and real-time data exchange between the simulator and controller was established using OPC DA. The fuzzy controller applies split-range control to two valves and was tested under 21 different initial pressure conditions using five distinct defuzzification strategies, resulting in a total of 105 unique test scenarios. System performance was evaluated using both error-based metrics (MSE, RMSE, MAE, IAE, ISE, ITAE) and dynamic response metrics, including overshoot, undershoot, rise time, fall time, settling time, and steady-state error. A web-based simulation interface was developed in Python using the Streamlit framework. Although demonstrated here for sour water treatment, the proposed fuzzy expert system is general-purpose. It can be adapted to other industrial processes by adjusting settings, rules, and input/output ranges based on expert knowledge or available documentation. The main contributions of this work include a reusable digital twin-based framework, fuzzy control for pressure regulation, and a detailed performance comparison of defuzzification methods.}

\keywords{Fuzzy Expert System, Fuzzy Inference System, Fuzzy Logic, Digital Twin, Unisim Design, Artificial Intelligence}



\maketitle

\section{Introduction}
\label{sec:introduction}

% Problem statement
% В начале должен быть HOOK! 
Oil is a key source of energy worldwide, as most vehicles, airplanes, engines, and other machinery rely on it. However, improperly processed oil can significantly harm the environment. Crude oil contains many acidic components, such as carbon dioxide ($\text{CO}_{2}$) \cite{carbon_dioxide_crude_oil, carbon_dioxide_foaming_oil}, hydrogen sulfide ($\text{H}_{2}\text{S}$) \cite{hydrogen_sulfide_oil_presence_two, hydrogen_sulfide_oil_presence}, and other organic acids \cite{crude_fuel_acids}. Carbon dioxide ($\text{CO}_{2}$) is often injected into oil reservoirs to enhance oil recovery, which increases production but also affects the final product composition and cost \cite{carbon_dioxide_crude_oil}. Further processing of the extracted oil results in the release of acidic components, particularly sour water. Sour water containing $\text{H}_{2}\text{S}$, $\text{C}\text{S}_{2}$, and $\text{C}\text{O}_{2}$ is discharged into the environment. Its gaseous vapors negatively impact the atmosphere, plants, soil, water, and other living organisms \cite{hydrogen_sulfide_oil_presence, acidic_gases_environment, acidic_water_environment_impact, petroleum}. Additionally, acidic flows passing through metal pipelines cause active corrosion, leading to significant damage and degradation of the metal \cite{metal_corrosion_acids, hydrogen_sufide_corrosion_mdpi}.

There are several approaches to categorizing oil quality, with experts primarily dividing it by sulfur content. Low-sulfur fuel contains less than 1\% sulfur \cite{fuel_types}, while high-sulfur fuel typically ranges between 1\% and 3.5\% \cite{fuel_types}. In oil and gas processing, any mistake at the facility — such as incorrect dosing during the extraction of acidic components ($\text{H}_{2}\text{S}$, $\text{CO}_{2}$), incorrect pressure calculations, or improper flow control — can damage the final product quality by affecting the sulfur composition \cite{sulfur_influence_cost}. This, in turn, can lead to significant financial losses and pose health risks to workers \cite{refining_influence, oil_splill_health}.

Purifying sour water is essential in modern life to remove or minimize the acids released during oil processing. First, reducing the discharge of sour water helps protect the environment. Second, water purified of acidic components can be reused in other technological processes \cite{purified_water_usage_industry, petroleum_and_environment}, reducing the load on additional purification systems, or used by people in regions with limited water resources, thereby decreasing the consumption of fresh water for daily needs. Third, reusing purified water can also reduce transportation and distribution costs.

% Main message
The current paper presents a novel fuzzy expert system for industrial applications, using the process of collecting and purifying sulfur water as an example, with a digital twin used for data collection. The fuzzy logic system, mimicking human reasoning, controls optimal split-range operation in a three-phase separator to maintain the desired pressure, the first step in water purification. The digital twin is built using Honeywell’s UniSim Design R492 release, accurately mirroring the real technological process.

% Significance
This research can significantly reduce the risk of harm to people by minimizing their involvement in the technological process and decreasing the release of acidic components into the environment. It enables the production of purified water that can be reused in other industrial processes or in daily human activities, as the final pH level ranges from 4.5 to 8.5. The fuzzy expert system allows non-expert (junior) workers to participate in the control process, as all rules and knowledge were developed based on the reviewed and verified technological process documentation. The digital twin of the entire technological process was also built from this documentation, reflecting a real process implemented at one of Kazakhstan’s oil and gas facilities in 2009. Details of the original documentation are withheld due to confidentiality and non-disclosure agreements.

% Contribution
The main contributions of this paper are as follows:
\begin{itemize}
    \item \textit{Implementation of a Digital Twin and Dataset Generation.} A digital twin model was developed in Unisim Design R492 release version to simulate the process of collecting and purifying sour water, enabling the generation of a custom dataset by measuring variable ranges and behaviors for rule implementation. This model is crucial for analyzing critical situations in the oil and gas industry, helping prevent incidents in which even minor errors could lead to fires or explosions, given the industry's highly flammable nature.
    \item \textit{Development of a Fuzzy Expert System with Split-Range Control.} A collection of fuzzy sets and control rules was applied to achieve higher quality control results. This fuzzy expert system can be adapted not only to the selected process but also to other industrial processes by learning and incorporating expert knowledge of new technological systems. The control rules are based on documented factory knowledge regarding variable dependencies and required input/output ranges. Fuzzy logic is used both for making control decisions and proactively maintaining system variables in safe states to prevent potential failures.
    \item Comprehensive Evaluation of Control System Performance. The fuzzy control system was tested under 21 different initial pressure conditions using five different defuzzification methods. Performance was evaluated using several metrics, including MSE (Mean Squared Error), RMSE (Root Mean Squared Error), MAE (Mean Absolute Error), IAE (Integral of Absolute Error), ISE (Integral of Squared Error), and ITAE (Integral of Time-weighted Absolute Error). Additionally, control performance was characterized by measuring overshoot, undershoot, rise time, fall time, settling time, and steady-state error to assess how quickly and accurately the system reaches the desired target values.
    \item \textit{Deployment of an Interactive Web-Based Simulation Tool.} A web application was developed using the Streamlit framework to provide a user-friendly interface for simulating the fuzzy split-range pressure control system. The tool allows users to configure input parameters, run simulations, visualize system behavior through interactive plots, and view key performance metrics. The application is publicly accessible at: https://temirbolat-fuzzy-split-range.streamlit.app. This contribution supports broader usability, reproducibility, and accessibility of the proposed approach.
\end{itemize}

The paper is organized as follows. Section~\ref{sec:introduction} provides the introduction. Section~\ref{sec:related_works} presents an overview of related work along with the theoretical background on expert systems. Section~\ref{sec:methods} describes the methods used to achieve the results. It includes key research details about the technological process (Section~\ref{sec:technological_process_description}), data collection through the digital twin (Section~\ref{sec:digital_twin}), system identification (Section~\ref{sec:system_identification}), data transfer (Section~\ref{sec:data_transfer}), fuzzy logic theory (Section~\ref{sec:fuzzy_logic}), and a step-by-step description of the proposed expert system implementation (Section~\ref{sec:proposed_approach}). Section~\ref{sec:experimental_results} presents the experimental results, the selected evaluation metrics, discusses encountered issues, and analyzes the outcomes. Finally, Section~\ref{sec:conclusion} concludes the paper by summarizing the work performed and highlighting the key findings of the research.

\section{Related works}
\label{sec:related_works}

% Начинается c вводного предложения.
% Пример: Есть очень много работ на эту тему ИЛИ Это новая область и работ мало ИЛИ Есть несколько лагерей исследователей ИЛИ Есть 2 подхода решить эту задачу и потом вы их рассказываете.
% Надо сделать так, что если переставить местами параграфы, то невозможно будет прочитать или логиски их связать, чтобы читал именно последовательно, а не выборочно. Должна быть история.
\subsection{Expert system theory}
According to \cite{expert_system_engineering_book}, artificial intelligence (AI) can be categorized into several major areas, including expert systems (ES), fuzzy logic (FL), genetic algorithms (GA), artificial neural networks (ANN), robotics, and others.

\begin{figure}[ht]
\centering
\includegraphics[width=0.8\textwidth]{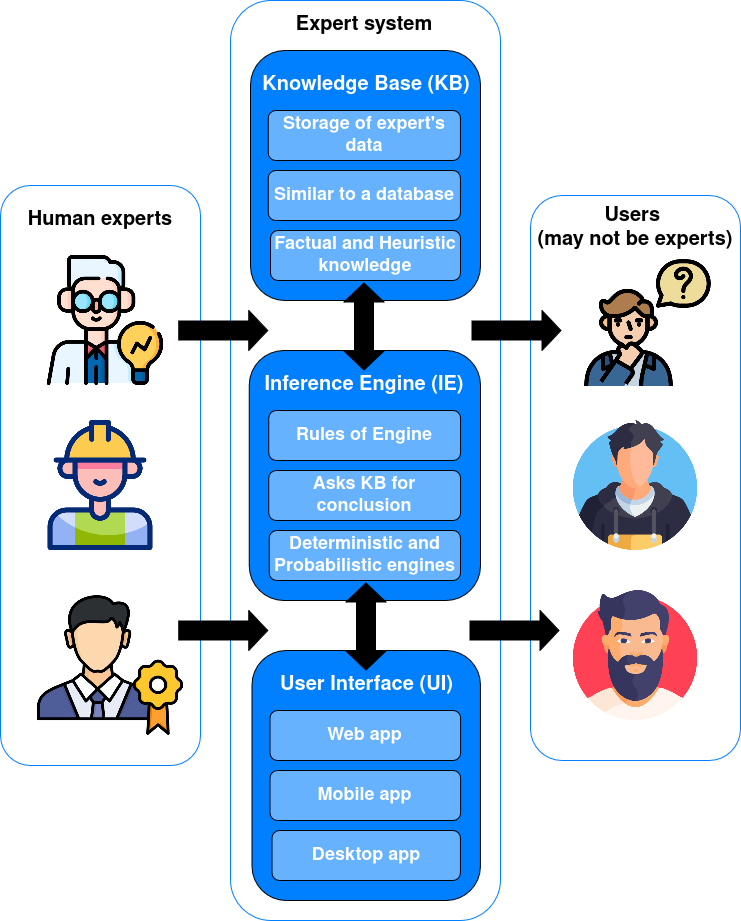}
\caption{Components of an Expert System}
\label{fig:expert_system_structure}
\end{figure}

Following that, in \cite{ai_overview_applications_power_electronics}, expert systems are described as one of the earliest AI methods to be effectively applied in various industries, including healthcare, finance, and engineering \cite{random_forest_expert_system}. An expert system \cite{ai_overview_applications_power_electronics, expertsystem_diognes_treatment, ai_techniques} is a branch of AI designed to support computer-based decision-making by utilizing a large amount of predefined information and rules adapted to specific failure scenarios. It operates based on expert-level knowledge \cite{svm_based_exp_system} and helps prevent critical issues such as diagnostic errors, degradation in product quality, or loss of control over key process parameters. Every expert system generally consists of three core components: the knowledge base (KB), the inference engine (IE), and the user interface (UI) \cite{expertsystem_diognes_treatment, ai_techniques, expertsystem_components_first, es_components_list_second}. Figure~\ref{fig:expert_system_structure} illustrates the main components of a typical expert system.

The knowledge base (KB) serves as the foundation of any expert system (ES) \cite{ie_kb_springer}. It consists of a predefined set of rules, process limitations, and logical statements that resemble the knowledge of human experts in a specific domain \cite{expertsystem_diognes_treatment, expert_systems_fundamentals_book}. The knowledge base is created and populated by a domain specialist, who defines the instructions using a formal language that the system can interpret and process.

The inference engine (IE) refers to the decision-making component of an expert system \cite{ie_kb_springer}. It applies the rules and operations defined in the knowledge base (KB) \cite{ai_techniques} and, after processing the input data, generates the corresponding output. The method by which the system derives conclusions from the knowledge base may vary depending on the reasoning strategy used.

The User Interface (UI) is the communication layer through which the user interacts with the expert system \cite{expert_system_kb_ie_ui_components}. This component can be implemented using various existing solutions, such as web interfaces, mobile applications, or desktop software, and can be developed in any widely used programming language.

There are various approaches to implementing expert systems. Still, the authors in \cite{expert_systems_types} identify five primary types: rule-based expert systems, frame-based expert systems, fuzzy expert systems, neural expert systems, and neuro-fuzzy expert systems. In this work, we use fuzzy logic to develop an expert system based on documentation prepared by domain experts that describes the real-world process of collecting and purifying sour water at an oil and gas facility.

\subsection{Overview of research}
\label{sec:literature_review}

Several studies focus on predicting fuel quality at different stages, and some expert systems have been developed for the oil and gas industry. The quality of the final oil product is primarily determined by its condition and precise chemical composition.

One approach to predicting oil quality is the application of Artificial Intelligence (AI) algorithms \cite{lstm_neural_network_oil, refined_oil_article, anfis_aquila_optimizer_forecasting} trained on relevant datasets. In \cite{lstm_neural_network_oil}, the authors proposed a decomposition-based ensemble model (EEMD) combined with a Long Short-Term Memory (LSTM) network to forecast oil quality. The method involves decomposing the data using EEMD and applying dynamic time warping to obtain stable functions for model training. In contrast, \cite{refined_oil_article} proposes a multiple linear regression approach enhanced by Principal Component Analysis (PCA), using features such as Free Fatty Acid (FFA) content, Iodine Value (IV), and Moisture Content (MOIST). Both approaches \cite{lstm_neural_network_oil, refined_oil_article} demonstrate promising accuracy, but lack evaluation metrics such as recall, precision, or ROC-AUC, which are essential for understanding detailed model behavior.

In \cite{anfis_aquila_optimizer_forecasting}, the researchers propose combining the Adaptive Neuro-Fuzzy Inference System (ANFIS) with the Aquila Optimizer (AO) to form the AO-ANFIS algorithm, which is trained on real data from two countries. The authors of \cite{anfis_aquila_optimizer_forecasting, lstm_neural_network_oil} compare their models against existing standalone algorithms, including Genetic Algorithms (GA), LSTM networks, and other machine learning approaches. Despite strong performance metrics, the AO-ANFIS model could benefit from comparison with a broader range of forecasting methods. In addition, incorporating further mutation techniques during the optimization process may improve its overall accuracy.

The works in \cite{fuzzy_split_range_biodiesel_temp_control, fuzzy_split_range_fermentation_process, neuro_fuzzy_split_range_valves} apply fuzzy logic in combination with split-range control to regulate the valves and maintain key parameters at desired setpoints. The authors of \cite{fuzzy_split_range_fermentation_process, fuzzy_split_range_biodiesel_temp_control} focus specifically on temperature control using fuzzy logic. In their earlier study \cite{fuzzy_split_range_fermentation_process}, they model the fermentation temperature control process using fuzzy PI and fuzzy PID regulators. By controlling two valves to mix fluids of different temperatures, they demonstrate reduced energy consumption compared to a classical PID controller, as shown in simulations performed in SIMULINK. Their later work \cite{fuzzy_split_range_biodiesel_temp_control} validates the approach experimentally using a real biodiesel reactor, where a fuzzy controller with three inputs achieved the best control performance. The authors of \cite{neuro_fuzzy_split_range_valves} propose a more advanced architecture that combines a fuzzy controller with neural network predictors for a hybrid valve system. This configuration accounts for nonlinearity near valve transition points and improves flow manipulation. The results show that this neuro-fuzzy system outperforms classical PID control in terms of stability.

Fuzzy expert systems (FES) are advantageous when decisions must be made under conditions of limited or uncertain data, especially in complex domains such as the oil and gas industry \cite{fuzzy_expert_system_oil_gas_fuzzy_journal, fuzzy_expert_system_for_scoring}. In \cite{fuzzy_expert_system_oil_gas_fuzzy_journal}, the authors developed an FES to assist in selecting the most appropriate enhanced oil recovery (EOR) method for non-fractured reservoirs. The system uses reservoir properties, including temperature, depth, viscosity, API gravity, and permeability as input parameters. Based on these inputs, more than 600 fuzzy rules were formulated to recommend one of four EOR methods: $\text{CO}_{2}$ injection, hydrocarbon gas injection, polymer flooding, or steam injection. It demonstrated higher accuracy than the Bayesian model, achieving an error rate of only 19\%, compared to 34\% for the Bayesian approach. However, the system was limited to only four predefined EOR methods and lacked adaptability, as it could not update or expand its rule base over time.

Similarly, the authors of \cite{fuzzy_expert_system_for_scoring} designed an FES to assess the safety of offshore oil platforms. The system evaluates overall risk by aggregating compliance scores from 14 operational areas across each platform. It was tested on 10 platforms and produced consistent results; however, in some cases, the final safety score was overly influenced by internal fuzzy logic rather than actual safety indicators, potentially leading to biased or misleading conclusions. In a different industrial context, the study presented in \cite{fuzzy_expert_system_palm_oil_exp_syst_appl} applied fuzzy logic to reduce oil loss during palm oil production. Fuzzy models were developed for each stage of the production process based on expert interviews from two factories in Indonesia. Moreover, the researchers implemented the fuzzy system in Microsoft Excel and MATLAB, and the Excel-based results closely matched those produced by MATLAB.

When considering expert systems in petroleum engineering, it becomes apparent that their number is relatively small, and they tend to focus on narrow, task-specific applications, as shown in \cite{ontology_petroleum, expert_syste_flow_correlation, expert_system_petroleum_production, expert_system_pollution}. In \cite{ontology_petroleum}, the authors developed a new Plant Ontology (PO) to integrate data from various sources for offshore petroleum plants, following established standards \cite{ontology_petroleum_digital_twin, ontology_iso}. The proposed O3PO model demonstrated strong predictive accuracy, achieving a Mean Absolute Percentage Error (MAPE) of 1.54\% and a Root Mean Square Error (RMSE) of 3.1\%. This tool introduces a new domain ontology to standardize entities in offshore petroleum production plants, supporting engineers and IT professionals, and it has been practically applied at a Brazilian offshore facility. Although the system provides a wide range of capabilities, it would require further extension to support sensor and actuator modeling.

Next, the articles \cite{expert_syste_flow_correlation} and \cite{expert_system_petroleum_production} describe the development of expert systems to improve petroleum industry operations by addressing specific challenges and proposing corresponding solutions. The first system focuses on selecting the most suitable multiphase flow correlation to predict pressure drops without requiring direct pressure measurements.
The second system is designed to automate monitoring, control, and diagnostics of petroleum production and separation processes, particularly in remote or hard-to-access locations. The expert system presented in \cite{expert_syste_flow_correlation} achieved a notable reduction in prediction error, with MAPE ranging from 0.01\% to 3\% in most cases. It utilized an extensive database of pressure drop values across different flow correlations, processed using VBA code. The monitoring and diagnostic system described in \cite{expert_system_petroleum_production} is intended to integrate real-time data analytics to predict potential equipment failures proactively.

The Structural Auto-Adaptive Intelligent Grey Model (SAIGM) was introduced in \cite{new_grey_model_forecasting} to forecast long-term petroleum consumption by determining optimal parameters. One key advantage of this model is that it does not require data pre-processing or filtering. Its innovation lies in structural flexibility and reduced reliance on expert modeling knowledge. However, a critical limitation of the study is that the prediction function was derived purely theoretically, without validation through practical experimentation, which highlights a potential weakness in its real-world applicability.

The integration of digital twin (DT) technology with artificial intelligence (AI) for control applications has rapidly advanced in recent years. Major Application Domains include:
\begin{itemize}
    \item \textbf{Manufacturing.} AI-enabled digital twins are driving closed-loop manufacturing, real-time process optimization, and dynamic system reconfiguration \cite{Mo2023A, Karkaria2025An, Alfaro-Viquez2025A}.
    \item \textbf{Smart Grids and Energy.} AI-driven DTs enhance predictive maintenance, asset management, and decision-making in complex energy systems \cite{Kobayashi2023Explainable, Apostolakis2023Digital}.
    
    \item \textbf{Healthcare.} Digital twins, powered by AI, enable personalized treatment, predictive analytics, and enhanced patient care through real-time data integration and simulation \cite{Tao2024Physical, Vallée2023Digital}.
    
    \item \textbf{Transportation}. DTs combined with AI control strategies optimize traffic management, vehicle suspension, and power converter control \cite{Huang2025Artificial, Fan2023Ubiquitous}.
    
\end{itemize}

% Нужно обозначить несколько проблем, не одну, а затем сказать, что в этой статье ты будешь заниматься именно вот такой проблемой
Although the reviewed approaches and expert systems demonstrate high accuracy in predicting oil quality and supporting operations through fuzzy logic and machine learning, they do not address the critical step of purifying sour water after crude oil processing, which can significantly affect environmental and operational outcomes. While many of these studies utilize data from real industrial facilities, they often do not account for variations in equipment types, configurations, and performance characteristics, which are essential for developing truly adaptable and robust control systems.

\section{Methods}
\label{sec:methods}

% \textcolor{red}{The current section consists of the following subsections: a description of the technological process, data collection, evaluation metrics, fuzzy logic, and the proposed approach for constructing the expert system, including its implementation in MATLAB.}

\begin{figure*}[ht]
\centering
\includegraphics[width=\textwidth]{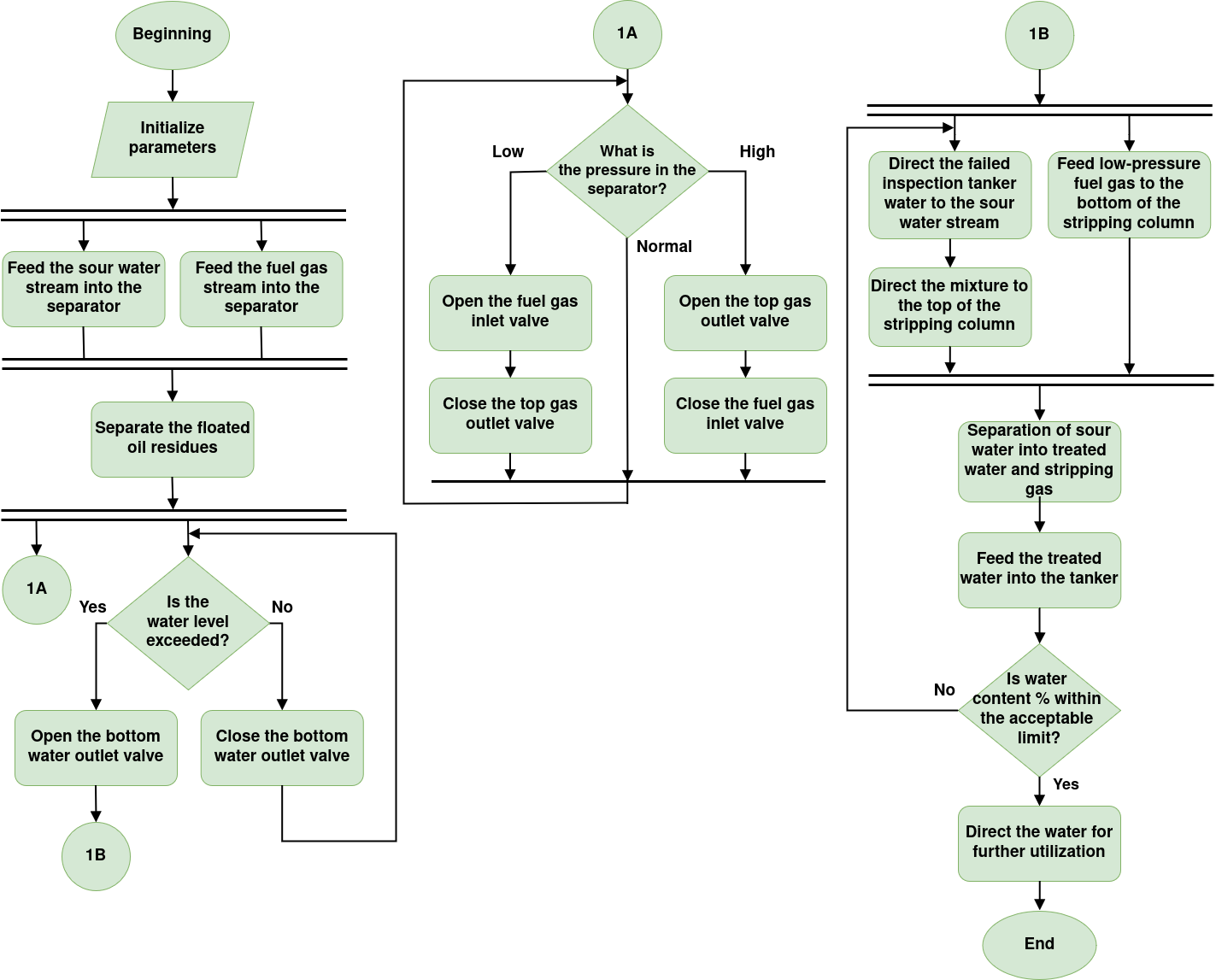}
\caption{Block diagram of the technological process}
\label{fig:block-diagram}
\end{figure*}

% \textcolor{red}{The Technological Process Description subsection provides information about the approach to collecting and purifying acidic water, the main process parameters that require control, a block diagram of the process, and a step-by-step explanation of each stage. The Data Collection section includes three subsections: Digital Twin Data, System Identification, and Data Transfer. The theoretical part of the digital twin subsection explains its applications, how it was built, and how it is used in this research to generate a custom dataset. The system identification subsection describes how a mathematical model was estimated to verify the rule logic based on test data from the valve, including the delay before connecting it to the digital twin. The data transfer subsection outlines how data is transmitted from the digital twin to the fuzzy inference system and back, and what approach was used to enable real-time communication. The Evaluation Metrics subsection lists the selected metrics used to evaluate the performance of the fuzzy split-range control and briefly describes the role and importance of each. Finally, the Proposed Approach subsection presents the core methodological steps of the developed expert system, followed by its integration with the custom digital twin and a summary of its performance results.}

\subsection{Technological process description}
\label{sec:technological_process_description}

In the oil and gas industry, the collection and treatment of sour water during crude gas distillation is a crucial technological process. This process consists of several control plants. Figure~\ref{fig:block-diagram} illustrates the block diagram of the whole process. The process mainly involves three central technological units: a three-phase horizontal separator, an absorber, and a storage tank.

Initially, the high-pressure fuel gas stream and the sour water stream containing liquid hydrocarbon impurities are fed into the three-phase separator. Due to density differences, the separated oil rises to the surface of the water, then overflows a partition and exits the separator through a pipeline. The sour water, now separated from hydrocarbons, is directed to the next unit—the absorber. The pressure inside the separator is maintained by two valves that regulate gas flow at the inlet and outlet of the unit. One of the valves controls the flow of high-pressure fuel gas into the separator, increasing the internal pressure. The second valve, installed on the outlet pipeline, reduces the pressure within the unit. The water level in the separator is monitored using a level gauge and controlled by the valve.

The sour water collected from the bottom of the main compartment flows through a valve into the top of the stripping column, which has 25 sieve trays. Inside the column, low-pressure fuel gas is added at the bottom. As the gas rises, it removes acidic components from the sour water. As the water moves down through the trays, it becomes cleaner. In the end, the process separates the flow into two parts: clean (treated) water and stripping gas. Clean water is sent to a tanker, while the stripping gas is sent to another stripping column. Since the focus here is on water treatment, the process for handling the stripping gas is not described.

The treated water flows into a horizontal storage tank where any remaining gases are collected and removed. The water level in the tank is measured by a level sensor and maintained by a control valve at the outlet. A pressure sensor monitors the separator pressure, and an ejector removes gas.

\subsection{Data Collection}
\label{sec:data_collection}

\subsubsection{Digital twin data}
\label{sec:digital_twin}

We performed data collection using the Digital Twin (DT) approach. In general, a digital twin is a virtual model that represents a real physical object or process \cite{digital_twin_intro_chapter, PHANDEN2021174}. Digital twins have a wide range of applications \cite{digital_twin_application}, including:
\begin{enumerate}
   \item To simulate how a real system works by using actual data and showing how it behaves in different situations;
    \item To collect data for detailed analysis or to train AI models;
    \item To check if the system is safe and stable before using it in real life.
\end{enumerate}

\begin{figure*}[tb]
\centering
\includegraphics[width=0.95\textwidth]{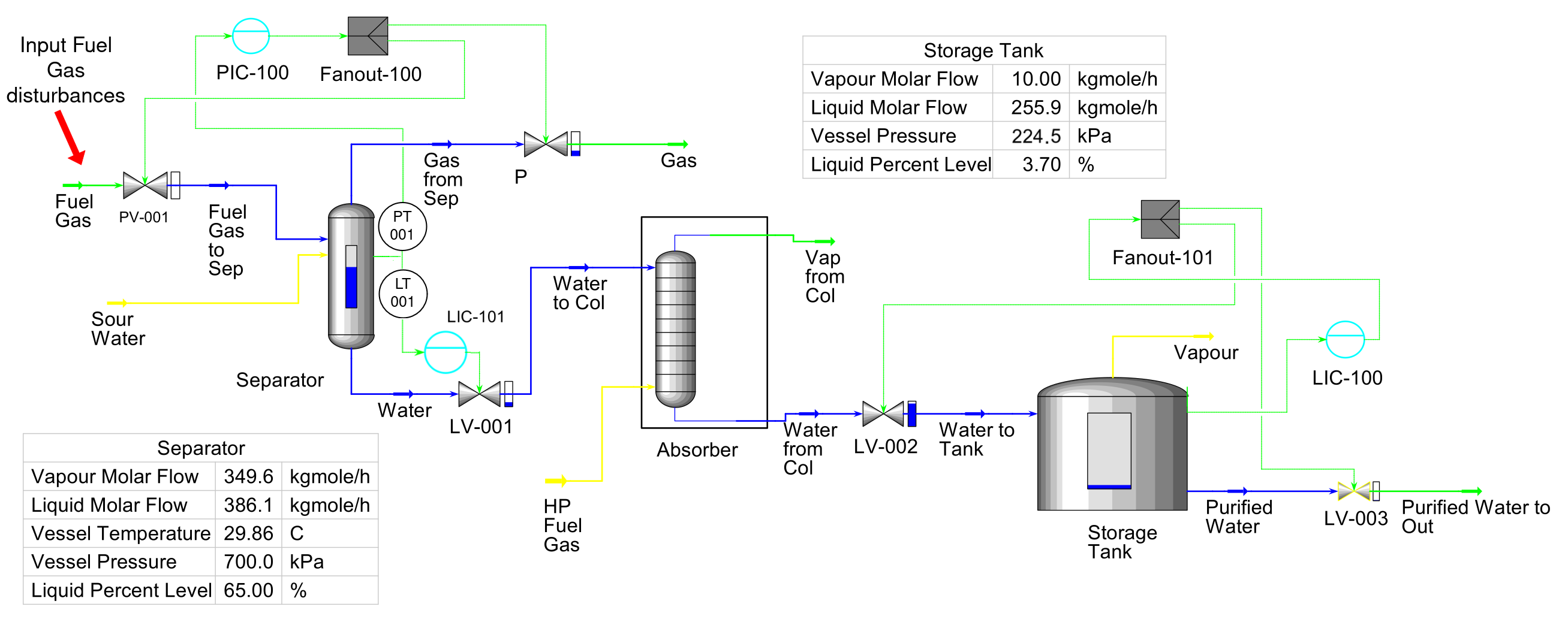}
\caption{Process flowsheet of the sour water treatment digital twin developed in UniSim Design R492 (Thermodynamic package: Sour Peng–Robinson).}
\label{fig:unisim_design_pfd}
\end{figure*}

In a different research, we created a digital twin of the sour water collection and purification process using the Unisim Design R492 release application developed by Honeywell. The software was selected because this process is standard in the oil and gas industry, where Unisim Design is widely used for simulation and modeling. Figure~\ref{fig:unisim_design_pfd} illustrates the complete UniSim Design digital twin with consistent equipment numbering and annotated callouts showing the control and measurement structure.

The digital twin comprises three main control units: the separator, absorber, and storage tank. It includes five valves, two of which (PV-001 and PV-002) are actively controlled to maintain the separator pressure under the supervision of the pressure controller PIC-100. The process disturbance originates from variations in the fuel-gas stream supplied to the system, which passes through the inlet control valve PV-001. Step changes in this flow are introduced to evaluate the pressure-control response of the separator. The pressure is measured by the pressure transmitter PT-001, which provides feedback to the controller for real-time pressure regulation, while the liquid level in the separator is monitored by the level transmitter LT-001. The pressure in the storage tank is 224.5 kPa, in accordance with the design specification of the sour-water buffer vessel.

The separator is also equipped with a level control loop, where the level indicator controller LIC-101 manipulates the drain valve LV-001 to stabilize the liquid level. 
The purified water collected in the storage tank is regulated by another control loop involving the LIC-100 controller and the LV-003 valve. 
Pressure transmitters and control valves are numbered consistently throughout the simulation flowsheet to ensure traceability between the model and its physical counterpart.

According to the documentation, pressure is measured in kilopascals (kPa), ranging from 0 to 1000 kPa (equivalent to 0 to 10 bar), while valve positions are expressed as percentages from 0\% (fully closed) to 100\% (fully open). 
As specified in the documentation, the optimal operating pressure for the three-phase separator lies between 400 and 600 kPa, with 500 kPa being ideal for achieving better compositional separation of the process stream. 
The digital twin operates in both static and dynamic modes, allowing real-time simulation and analysis with adjustable parameters. 
The Sour-Peng–Robinson thermodynamic package was used to combine all necessary thermodynamic properties and component data for phase equilibrium calculations. 
The raw gas inlet stream modeled in the system consists of the following components: methane ($\text{CH}_{4}$), ethane ($\text{C}_{2}\text{H}_{6}$), propane ($\text{C}_{3}\text{H}_{8}$), carbon dioxide ($\text{CO}_{2}$), water ($\text{H}_{2}\text{O}$), isobutane ($\text{C}_{4}\text{H}_{10}$), isopentane ($\text{C}_{5}\text{H}_{12}$), hydrogen sulfide ($\text{H}_{2}\text{S}$), and nitrogen ($\text{N}_{2}$).

\subsubsection{System identification}
\label{sec:system_identification}

Before connecting the digital twin to the fuzzy logic controller in MATLAB, we generated valve position data and performed system identification to build a transfer function. This helped us validate how the fuzzy controller affects valve behavior.

\begin{figure}[ht]
\centering
\includegraphics[width=0.9\textwidth]{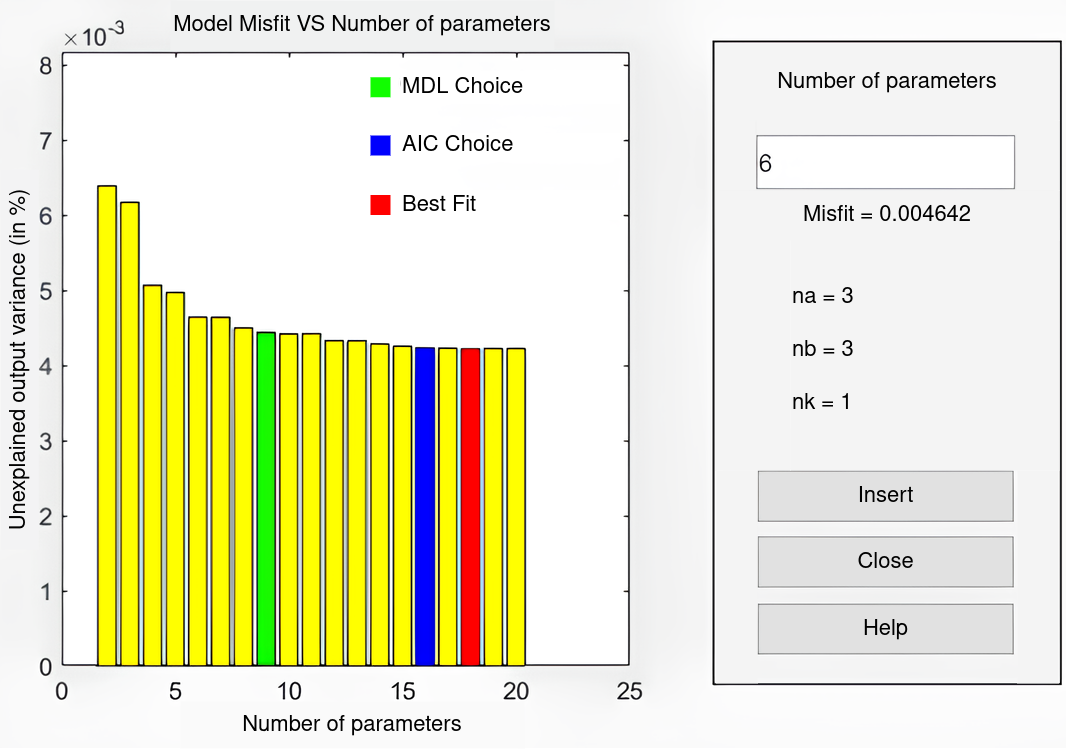}
\caption{Model's misfit based on poles and zeros pairs. The misfit metric quantifies how well the model output matches the measured data; lower values indicate a better fit.}
\label{fig:arx_model_figure}
\end{figure}

The data was generated using Python. In total, the script produced 1000 records with a time step of 0.5 seconds, resulting in a 500-second simulation. The control input signal was formed by combining three components: a smooth sinusoidal wave (to represent gradual changes), random noise (to simulate signal uncertainty), and step changes (to model sudden control actions). Input and output values ranged from 0 to 1, corresponding to 0 \% and 100 \% valve positions. To simulate real-world valve dynamics, a 0.5-second delay was added, which is possible for pneumatic control valves \cite{valve_pneumatic_delays}. Slight Gaussian noise was also added to the output to represent sensor inaccuracies. The final dataset included three columns: time (in seconds), input signal, and output signal. It was saved as a CSV file and imported into MATLAB. This data was then used in the System Identification Toolbox to create a dynamic model of the valve.

System identification is the process of creating a mathematical model of a system based on data collected from that system \cite{system_identification_book}. A mathematical model is a set of equations, formulas, or statements to represent a real-world system and its behavior \cite{mathematical_model_definition}. Using MATLAB's System Identification Toolbox \cite{matlab_system_identification_toolbox}, we built a transfer function for the valve using the earlier-generated CSV data. A transfer function is the ratio of a system’s output to its input, expressed in the Laplace "s" domain \cite{transfer_function_definition}. System identification helps engineers analyze complex systems, typically categorized as Multiple Input Multiple Output (MIMO), Multiple Input Single Output (MISO), or Single Input Single Output (SISO). In MIMO systems, multiple inputs influence multiple outputs. In MISO systems, several inputs affect a single output. In SISO systems, one input affects one output. In general, the system where we control the separator’s pressure by adjusting two valves is considered a MISO system, since multiple inputs (the two valves) influence a single output (the pressure). However, in this study, we used system identification only to get the valve’s transfer function. A typical form of a transfer function with some delay is the following:

\begin{equation}
G(s) = \frac{Y(s)}{U(s)} = \frac{b_0 s^m + b_1 s^{m-1} + \cdots + b_m}{a_0 s^n + a_1 s^{n-1} + \cdots + a_n} \cdot e^{-sT_d}
\label{eq:transfer_function}
\end{equation}
\noindent where,
\begin{itemize}
  \item $Y(s)$ is the output signal in the Laplace domain;
  \item $U(s)$ is the input signal in the Laplace domain;
  \item $b_i$ is the $i$-th coefficient of the numerator;
  \item $a_i$ is the $i$-th coefficient of the denominator;
  \item $s$ is the Laplace variable;
  \item $T_d$ is the time delay (in seconds).
\end{itemize}

In MATLAB, we imported the valve data in the time domain, specifying a sample time of 0.5 seconds. A total of 1000 samples were loaded. Before developing the model, the data were preprocessed by removing the mean from both the input and output signals. The resulting dataset was then divided into two parts: working data and validation data—the first 500 samples served as the working dataset, and the remaining 500 as the validation dataset. The validation data was used to test the accuracy of the estimated models. To assess the input delay and the number of poles and zeros for a transfer function, we constructed a SISO Auto-Regressive with eXogenous input (ARX) model. The ARX model is a mathematical model that predicts the system's output from its past inputs and outputs \cite{arx_model_definition}. Mathematically, the ARX model for a SISO system can be expressed as follows:

\begin{equation}
\begin{aligned}
&y(t) + a_1 y(t - 1) + \dots + a_{n_a} y(t - n_a) = \\
&\quad b_1 u(t - n_k) + \dots + b_{n_b} u(t - n_k - n_b + 1) + e(t)
\end{aligned}
\label{eq:arx}
\end{equation}
\noindent where,
\begin{itemize}
  \item $y(t)$ is the output of the system at time $t$;
  \item $u(t)$ is the input at time $t$;
  \item $a_i$ is the $i$-th coefficient related to past outputs;
  \item $b_i$ is the $i$-th coefficient related to past inputs;
  \item $n_a$ is the number of poles;
  \item $n_b$ is the number of zeros plus one;
  \item $n_k$ is the input delay;
  \item $e(t)$ is a white noise error term.
\end{itemize}

We tested all combinations of $n_a$, $n_b$ and $n_k$ with values from 1 to 10. After comparing the results, we chose the best combination. Figure~\ref{fig:arx_model_figure} demonstrates the model's misfit in \% based on all pole-zero pairs.

Using the best-fitting model parameters, including the corresponding number of poles, zeros, and input delay, we estimated the transfer function. The final transfer function for the valve is presented below:
\begin{equation}
G(s) = \frac{0.4455 s^2 - 1.14 \times 10^{-5} s + 0.003544}{s^3 + 0.447 s^2 + 0.007935 s + 0.003547}
\label{eq:transfer_function_substitued}
\end{equation}

\begin{figure*}[ht]
\centering
\includegraphics[width=\textwidth]{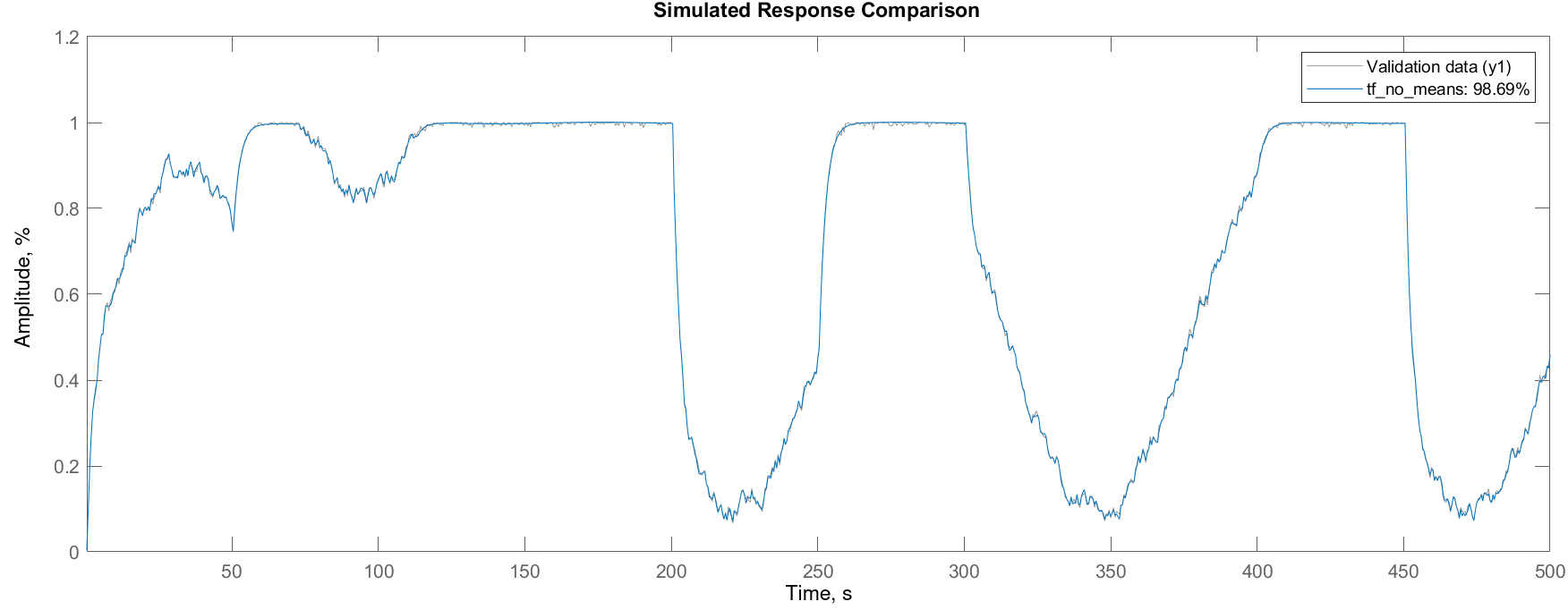}
\caption{Accuracy of the valve's transfer function model }
\label{fig:valve_transfer_function}
\end{figure*}

The transfer function is of third order because it has three poles. The transfer function's accuracy is 98.69\%. Figure~\ref{fig:valve_transfer_function} illustrates the accuracy of the model in comparison with the validation data.

\subsubsection{Data transfer}
\label{sec:data_transfer}

The current subsection describes the data transmission process from the Unisim Design application to MATLAB, which is essential for evaluating control accuracy. OPC (Open Platform Communications) is a widely adopted and reliable framework for data exchange between industrial equipment from various vendors. It involves several standards, including OPC DA (Data Access), OPC AE (Alarms \& Events), OPC Batch, OPC DX (Data eXchange), OPC HDA (Historical Data Access), OPC Security, OPC XML-DA (XML Data Access), and OPC UA (Unified Architecture) \cite{opc_mdpi}. Most OPC standards are based on the Distributed Component Object Model (DCOM) architecture, developed by Microsoft \cite{opc_dcom}.

Each OPC standard serves a specific purpose. For instance, OPC DA is primarily used for transmitting real-time data between systems such as SCADA (Supervisory Control and Data Acquisition), PLC (Programmable Logic Controller), DCS (Distributed Control System), and so on. OPC AE is used to send alarms, show messages, and record what the operator does \cite{opc_ae_elsevier}. OPC DX enables data exchange between OPC servers from different manufacturers over an Ethernet network. While OPC DA handles real-time data that continuously changes, OPC HDA is used to access stored data, which is useful when you need to view both current and past values. OPC Security defines mechanisms for access control and secure communication between clients and servers. OPC XML-DA and OPC UA can be used on different systems because they don’t depend on one specific platform. The former concerns web services that transmit data via SOAP (Simple Object Access Protocol) or HTTP (Hypertext Transfer Protocol) \cite{opc_xml_da_elseveir}. At the same time, the latter is not based on Microsoft COM technology and offers full cross-platform compatibility \cite{opc_ua_elsevier}.

We chose OPC DA for data transfer. OPC DA is a standard used in industrial automation to exchange real-time data \cite{opc_da_iee}. It consists of two main components: an OPC DA Server and an OPC DA Client. The server stores values at specific addresses and allows them to be read or updated, whereas the client connects to the server to read these values or send new ones.

\begin{figure}[ht]
\centering
\includegraphics[width=\columnwidth]{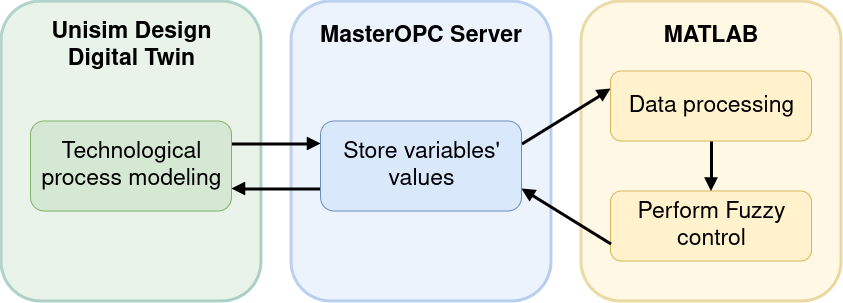}
\caption{Real-time data collection and transfer approach}
\label{fig:unisim_opc_matlab_big_cards}
\end{figure}

According to the Figure~\ref{fig:unisim_opc_matlab_big_cards}, both Unisim Design and MATLAB act as OPC Clients. Since OPC Clients cannot exchange data directly, an OPC Server is required to act as a bridge. For this purpose, the MasterOPC software was selected. The selected variables are sent from Unisim Design to the MasterOPC Server, with each variable configured by address, data type, and access mode. MATLAB then receives this data and, if needed, performs preprocessing to ensure values are within valid ranges for the fuzzification process. After fuzzy block control, the signals are sent back through the server to change the actuator values and complete the control loop. During the real-time simulation, both UniSim Design and MATLAB Simulink operated with a sampling period of 1 s. The OPC interface in UniSim (via the Master OPC Server) used a group-scan time of 1 s, consistent with the UniSim real-time integrator step. This configuration provided stable and realistic data exchange for the dynamic pressure-control process, which operates on a timescale of several seconds.

\subsection{Fuzzy Logic}
\label{sec:fuzzy_logic}

This subsection provides an overview of the key terms from fuzzy set and logic theory used in our research.

\subsubsection{Membership Functions and Fuzzy Sets}

A fuzzy set is a collection of elements that have varying degrees of membership, ranging from 0 to 1. In a universe of discourse $X$, a fuzzy set $A$ is defined by a membership function $\mu_A(x)$, which assigns to each element $x$ in $X$ a real number from the interval $[0, 1]$, indicating the degree to which $x$ belongs to $A$ \cite{Zadeh1965}, \cite{Zadeh1975}. This relationship can be described mathematically as:
\begin{equation}
\mu_A : X \rightarrow [0, 1]
\label{eq:membership}
\end{equation}
where $\mu_A(x)$ represents the degree of membership of $x$ in $A$.

\begin{figure}[htbp]
    \centering
    \begin{subfigure}[b]{\textwidth}
        \includegraphics[width=\textwidth]{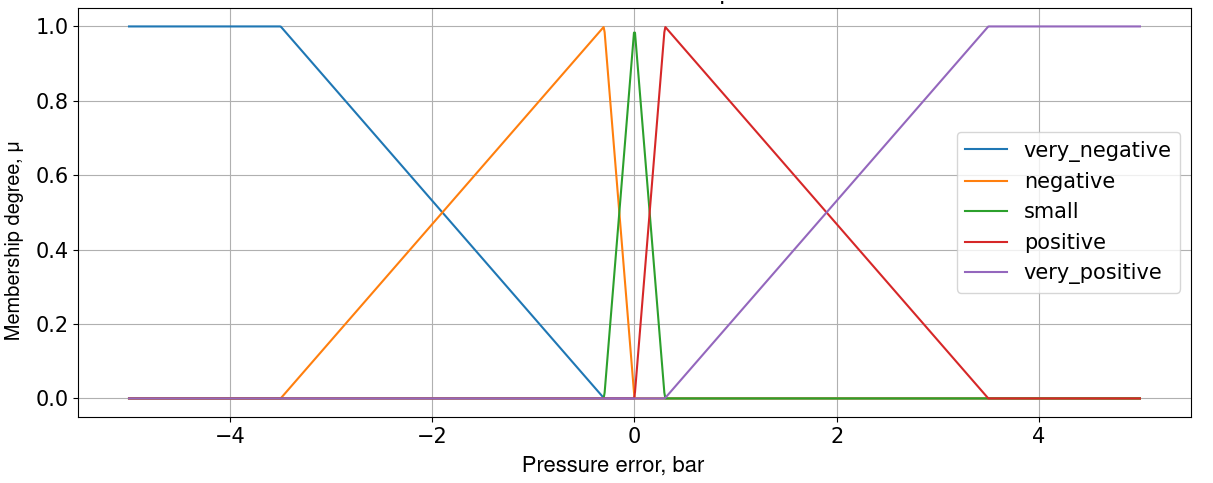}
        \caption{Pressure error membership functions}
    \end{subfigure}
    \hfill
    \begin{subfigure}[b]{\textwidth}
        \includegraphics[width=\textwidth]{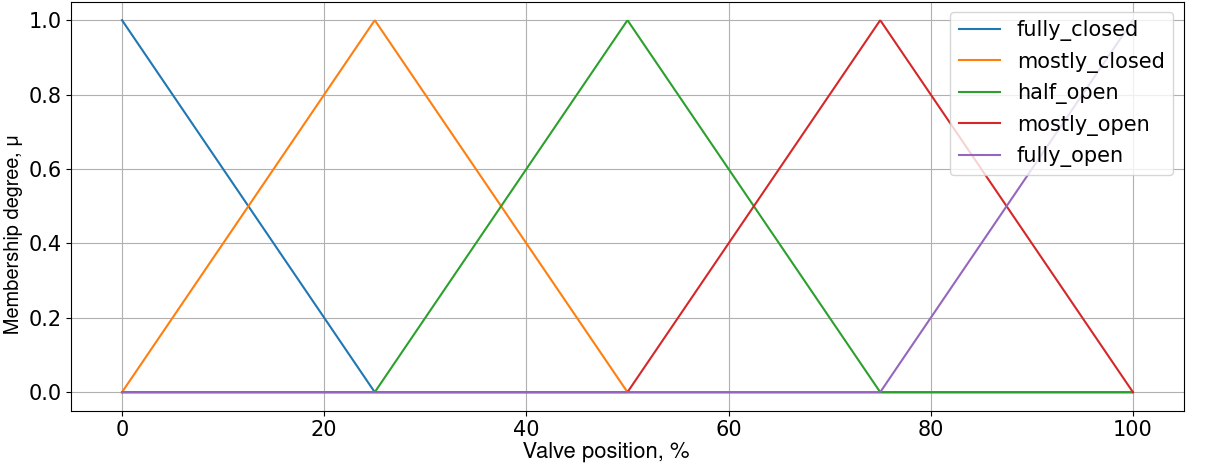}
        \caption{Input fuel/outlet gas valve membership functions}
    \end{subfigure}
    \caption{Fuzzy input and output membership functions}
\label{fig:fuzzy_membership_functions}
\end{figure}

Figure~\ref{fig:fuzzy_membership_functions} shows the membership functions for the input variable (PT-001) as well as for the output variables (positions of the PV-001 and PV-002). The plot for the outlet gas valve (PV-002) was not included, as its behavior is identical to the plot of the fuel gas valve (PV-001). Figure~\ref{table:fuzzy_sets_separator_control} stores information about fuzzy sets and their variable domains.

\begin{table}[tb]
\caption{Fuzzy attributes of the fuzzy split range separator's pressure control.}
\label{table:fuzzy_sets_separator_control}
\centering
\begin{tabular*}{\textwidth}{@{\extracolsep\fill}
>{\centering\arraybackslash}p{2.7cm}
>{\centering\arraybackslash}p{6.5cm}
>{\centering\arraybackslash}p{2.7cm}}
\toprule%
\textbf{Fuzzy variable} & \textbf{Term set} & \textbf{Domain} \\
\midrule
Pressure error, bar & T = \{Very negative, Negative, Small, Positive, Very positive\} & X = {[}-5, 5{]} \\
Input fuel valve, \% & T = \{Fully closed, Mostly closed, Half open, Mostly open, Fully open\} & X = {[}0, 100{]} \\
Outlet gas valve, \% & T = \{Fully closed, Mostly closed, Half open, Mostly open, Fully open\} & X = {[}0, 100{]} \\
\botrule
\end{tabular*}
\end{table}

\subsubsection{Fuzzy Operations}
The $\alpha$-cut (or Alpha cut) of a fuzzy set is a crisp set that contains all elements of the fuzzy subset $f$ whose membership values are greater than or equal to a specified threshold $\alpha$, where $0 < \alpha \leq 1$ \cite{Zadeh1965}, \cite{Zadeh1975}. Formally, it is defined as:
\begin{equation}
f_{\alpha} = \{ x \in X \mid \mu_f(x) \geq \alpha \}
\label{eq:alpha_cut}
\end{equation}
The $\alpha$-cut represents a way to approximate or discretize a fuzzy set into a conventional set, where all elements in the set have at least a degree of membership of $\alpha$.

To relate $\alpha$-cuts to standard set operations, consider two fuzzy sets $A$ and $B$. The union and intersection of these sets at the level of their $\alpha$-cuts are given by:
\begin{equation}
(A \cup B)_{\alpha} = A_{\alpha} \cup B_{\alpha}
\label{eq:alpha_union}
\end{equation}
\begin{equation}
(A \cap B)_{\alpha} = A_{\alpha} \cap B_{\alpha}
\label{eq:alpha_intersection}
\end{equation}
This means the $\alpha$-cut of the union of two fuzzy sets is the union of their respective $\alpha$-cuts, and similarly, the $\alpha$-cut of the intersection of two fuzzy sets is the intersection of their $\alpha$-cuts.

Furthermore, $\alpha$-cuts provide a helpful tool for simplifying complex fuzzy set operations. By selecting various values of $\alpha$, one can analyze different "layers" of the fuzzy set and understand how it behaves under different degrees of membership. This concept is crucial in applications such as decision-making and pattern recognition, where interpreting fuzzy sets as crisp sets via $\alpha$-cuts enables practical outcomes from imprecise data.

\subsubsection{Fuzzy Rules}

Fuzzy rules are an essential component of fuzzy logic, a branch of multi-valued logic that stems from fuzzy set theory and is used to handle approximate reasoning. These rules describe the relationship between a system's inputs and outputs and are designed to resemble the way humans make decisions. Fuzzy rules are commonly structured in an "IF-THEN" format, such as:
\begin{equation}
\text{IF } x \text{ is } A \text{ THEN } y \text{ is } B,
\label{eq:fuzzy_rule}
\end{equation}
where $x$ represents an input variable, and $A$ and $B$ are fuzzy sets defined over the input and output universes of discourse, respectively \cite{Zadeh1965}, \cite{Zadeh1975}. For instance, a fuzzy rule might state, "IF pressure is high THEN valve must be opened." This indicates that when the system's failure rate falls into the fuzzy set "high," the corresponding output (priority) also belongs to the fuzzy set "high."

Fuzzy rules are foundational in fuzzy inference systems, where the rule base consists of a set of such IF-THEN rules. These rules collectively allow the system to process vague, uncertain, or imprecise input information and generate reasonable outputs. The inference mechanism in these systems uses the rules to derive conclusions, often by applying a process known as fuzzy implication or aggregation.

One of the key strengths of fuzzy rules is their ability to model complex, nonlinear relationships flexibly. By mimicking human reasoning, they are handy in domains where decisions are made under uncertainty, such as control systems, decision-making, and expert systems. Moreover, fuzzy rules can easily accommodate multiple inputs and outputs, thereby expanding their applicability across fields such as robotics, medical diagnosis, and industrial control systems.

In summary, fuzzy rules enable systems to handle imprecision and ambiguity by modeling relationships between inputs and outputs that mirror human thought processes. This flexibility and adaptability make them crucial for solving problems that involve uncertainty and approximate reasoning.

\subsubsection{Fuzzy control system}

\begin{figure}[ht]
    \centering
    \begin{subfigure}[b]{0.49\textwidth}
        \includegraphics[width=\textwidth]{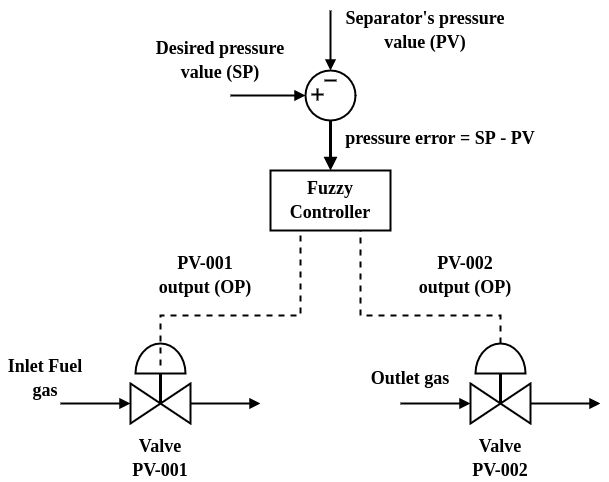}
        \caption{\centering Fuzzy split-range pressure control scheme}
    \end{subfigure}
    \hfill
    \begin{subfigure}[b]{0.49\textwidth}
        \includegraphics[width=\textwidth]{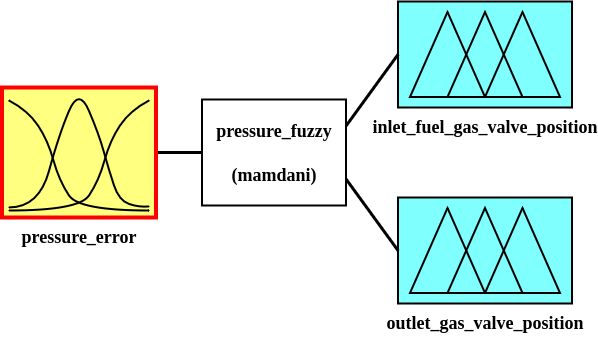}
        \caption{\centering Mamdani-based fuzzy pressure mapping}
    \end{subfigure}
    \caption{Fuzzy split-range control of separator pressure}
\label{fig:split_range_fuzzy_control}
\end{figure}

The current subsection describes the application of fuzzy logic in control and its implementation in MATLAB. Figure~\ref{fig:split_range_fuzzy_control} illustrates how the fuzzy controller processes the input signal and which parameters it regulates. The fuzzy logic controller adjusts the valve positions based on the separator's internal pressure (PT-001). The desired pressure value is defined during the fuzzification step, where appropriate ranges are assigned to each membership function. According to the documentation from the actual industrial process, the target pressure must be maintained within the range of 4 to 6 bar. At the output of the fuzzy controller, two signals are generated: the first corresponds to the required position of the PV-001 valve, which regulates the fuel gas input to increase pressure; the second corresponds to the required position of the PV-002 valve, which releases excess gas to reduce pressure. Their values range from 0 \% to 100 \%, where 0 \% represents a fully closed valve and 100 \% represents a fully open valve.

\begin{figure*}[ht]
\centering
\includegraphics[width=\textwidth]{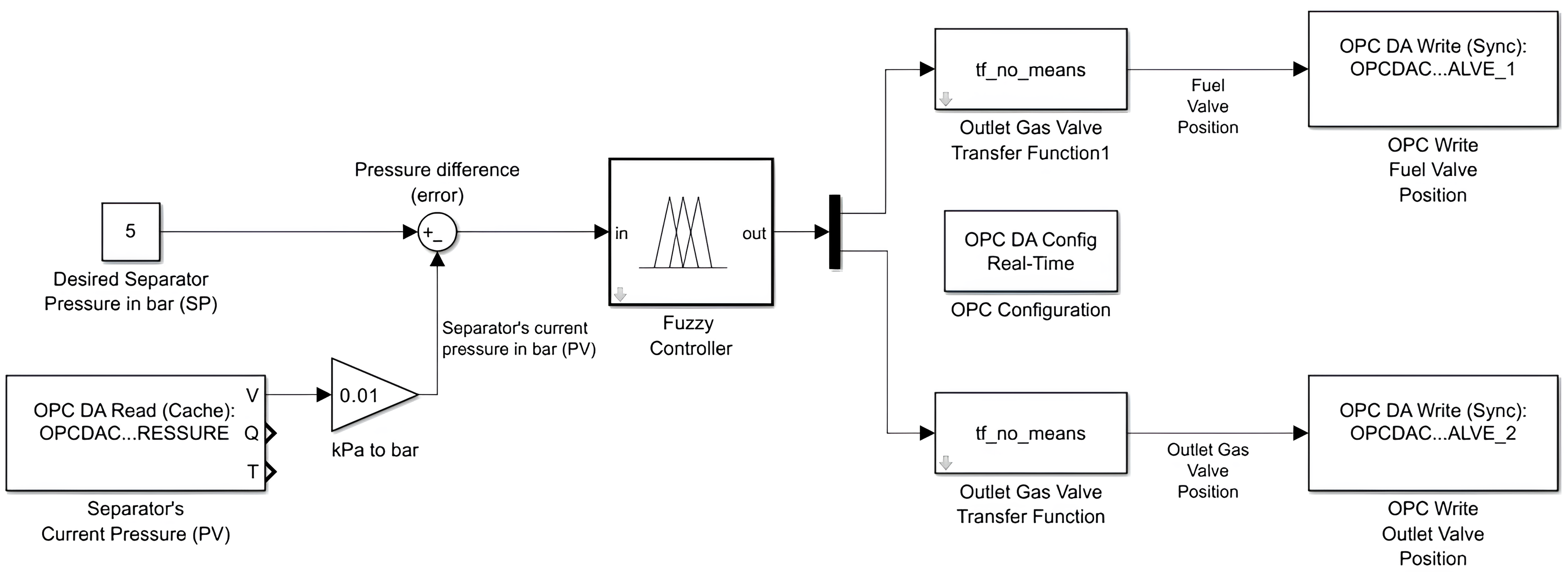}
\caption{Fuzzy split-range control in MATLAB SIMULINK}
\label{fig:matlab_simulink}
\end{figure*}

Figure~\ref{fig:matlab_simulink} illustrates the implementation of fuzzy control in MATLAB Simulink \cite{matlab_simulink}, where data is exchanged with the digital twin via an OPC server. Once the connection with the OPC Server is established, the incoming data is first processed to convert pressure from kPa to bar by multiplying it by a constant factor of 0.01, as the fuzzification ranges are defined in bar units. As the signal exits the fuzzy controller, it is split into two individual signals using a Demux block. Each signal is then directed to the corresponding valve's transfer function, which was obtained through system identification. Since the output after defuzzification lies between 0 and 1, but the digital twin expects a value between 0 and 100 \%, it is multiplied by 100 before being sent to the OPC server in real time, from which Unisim Design retrieves the value.

% TABLE BEGINNING
It is necessary to maintain the desired pressure in a three-phase separator, as specified in the process documentation, to ensure the appropriate composition of the output materials and to reduce the concentration of acidic components. This is achieved by manipulating two control valves. Table~\ref{table:rules_separator_pressure} shows the Fuzzy Rules used to control the pressure in the separator. A key point is that one input affects two outputs (refers to split-range control). In this case, the separator's pressure (PT-001) influences both the input fuel valve PV-001, which raises the pressure, and the output gas valve PV-002, which lowers the pressure.
\begin{table}[ht]
\caption{Fuzzy rules composed by the experts for the separator's pressure control}
\label{table:rules_separator_pressure}
\begin{tabular*}{\textwidth}{@{\extracolsep\fill}lcccccc}
\toprule%
& \multicolumn{1}{@{}c@{}}{Inputs} & \multicolumn{2}{@{}c@{}}{Outputs} \\ \cmidrule{2-2}\cmidrule{3-4}%
Rules & Pressure error & Input fuel gas valve & Outlet gas valve \\
\midrule
Rule 1 & Very positive & Fully open & Fully closed \\
Rule 2 & Positive & Mostly open & Fully closed \\
Rule 3 & Almost absent & Fully closed & Mostly closed \\
Rule 4 & Negative & Fully closed & Mostly open \\
Rule 5 & Very negative & Fully closed & Fully open \\
\botrule
\end{tabular*}
\end{table}

According to Table~\ref{table:rules_separator_pressure}, five fuzzy rules are used to control the separator’s pressure based on its error (the difference between the desired value and the current value). These rules are: "Very positive", "Positive", "Almost absent", "Negative", and "Very negative". The three-phase separator has two input streams (sour water and fuel gas) and two output streams (outlet gas and water), so the molar flows at the inputs and outputs must be balanced. When the pressure error is "Very negative", it means that the current pressure is much higher than the desired value. In this case, the input valve is fully closed to prevent further pressure increase, while the output valve is fully opened to release the excess pressure, quickly reducing the error. At the "Negative" level, the error is closer to zero, meaning the pressure is slightly above the desired value. In this case, the input valve remains closed, while the output valve is primarily open to achieve a smoother pressure reduction. In the "Almost absent" error state, where the pressure error is negligible and near zero, the fuel gas valve is fully closed.
In contrast, the outlet gas valve is mostly closed, since the outlet gas flow is lower than the input flow to the separator. If the error enters the "Positive" state, it means the current pressure is lower than the desired value, but not significantly. In this case, the input valve is mostly open to gradually increase the pressure, while the output valve remains closed. Finally, when the error reaches the "Very positive" state, the input valve is fully opened to restore the pressure rapidly. In contrast, the output valve remains fully closed to prevent any further loss.

\subsection{Proposed Approach}
\label{sec:proposed_approach}

We propose developing a fuzzy expert system based on data generated by a custom-built digital twin constructed from real technological process documentation from an industrial facility. The planned methodology is as follows:
\begin{enumerate}
    \item Construct a digital twin based on real equipment specifications to collect the necessary dataset;
    \item Perform system identification to model valve behavior using a linear approximation;
    \item Develop an expert system in MATLAB/SIMULINK based on the collected dataset, using a fuzzy logic approach under 21 different initial separator pressure conditions and five different defuzzification methods;
    \item Enable real-time data exchange between the digital twin and the fuzzy inference system via OPC;
    \item Compare the performance of the five defuzzification methods using calculated evaluation metrics and select the most effective approach;
    \item Implement a web interface to have a visual communication between the expert system and factory engineers.
\end{enumerate}

We will create a custom dataset using a digital twin built in Honeywell’s UniSim Design R492, which will closely replicate the real behavior of an industrial process based on official documentation from a Kazakhstani oil and gas facility focused on sour water collection and purification. We will model valve dynamics through system identification in MATLAB, using simplified linear approximations to represent valve behavior. To enable real-time communication between the simulator and the fuzzy controller, we will establish data exchange via the OPC DA protocol (MasterOPC app). We plan to design a fuzzy expert system that applies split-range control to manage two valves. We will implement the system in two environments: MATLAB/SIMULINK for real-time control and communication with the simulator, and Python for simplified plotting, CSV result extraction, and performance analysis under different defuzzification strategies. We will test the system under 21 different initial pressure conditions and apply five distinct defuzzification methods, resulting in 105 unique test scenarios. To evaluate the system’s performance, we will use a combination of error-based metrics (MSE, RMSE, MAE, IAE, ISE, ITAE) and dynamic response metrics, including overshoot, undershoot, rise time, fall time, settling time, and steady-state error.

\begin{figure*}[tb]
\centering
\includegraphics[width=\textwidth]{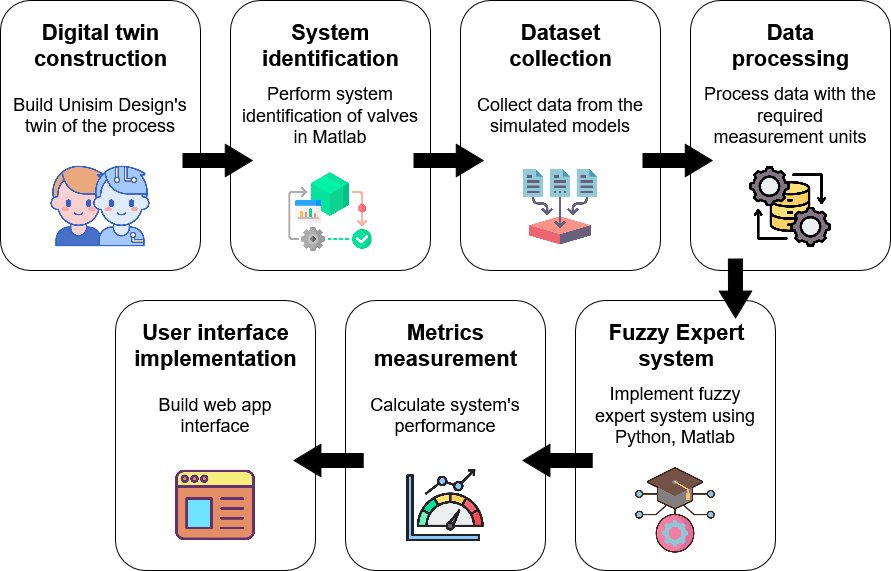}
\caption{Seven steps approach for expert system}
\label{fig:methodology_figure}
\end{figure*}

This section describes the steps for implementing an expert system for the process of collecting and purifying sulfur water using a digital twin. Figure~\ref{fig:methodology_figure} illustrates the sequential process of building the expert system.

\begin{figure*}[tb]
\centering
\includegraphics[width=\textwidth]{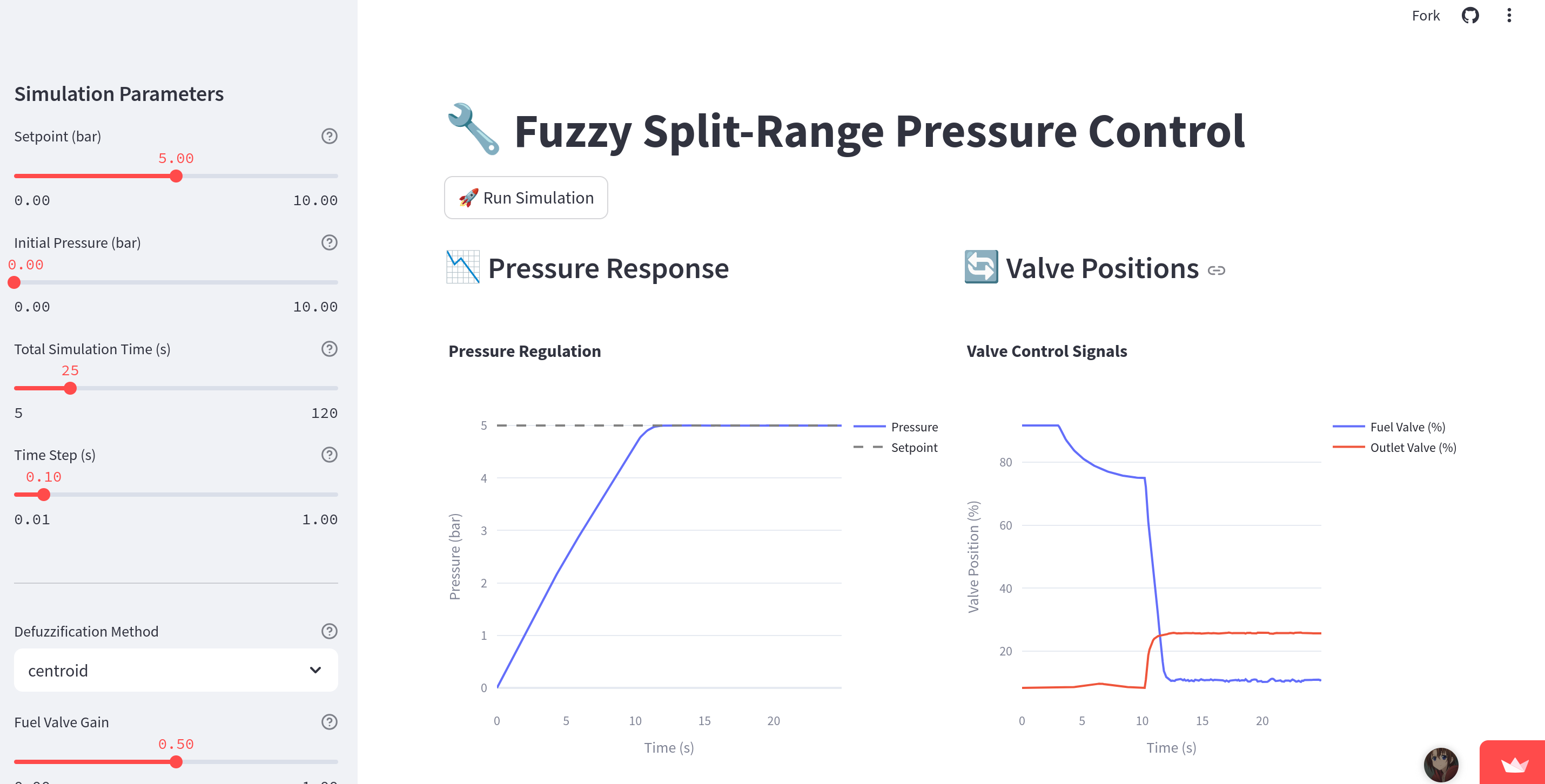}
\caption{Fuzzy split-range control web interface showing simulation settings and output plots for pressure regulation and valve positions}
\label{fig:streamlit-web-app}
\end{figure*}

According to the figure, the proposed approach consists of seven main steps, starting with the construction of the digital twin and ending with the implementation of the user interface. Section~\ref{sec:data_collection} provides detailed information about the digital twin construction and system identification steps. An overview of the remaining five stages in the proposed workflow is provided below:

\begin{itemize}
    \item \textit{Data collection.} Data collection is the step where data is transmitted from the digital twin to the fuzzy system in real time. In this setup, both the digital twin and the fuzzy system act as clients, requiring a server to store the transmitted values in real time. For this purpose, OPC DA was chosen as a bridge between Unisim Design (the digital twin) and the fuzzy control system (MATLAB). In this case, data on separator pressure, as well as input and output valve positions, is collected and saved in a file. Subsection~\ref{sec:data_transfer} provides detailed information about the data collection process.
    \item \textit{Data processing.} In this step, some data, such as the three-phase separator’s pressure, is received by the OPC server in kilopascals (kPa). In contrast, the technological process documentation specifies the pressure in bars. Therefore, the fuzzy logic system was designed to operate with pressure values in bars. Before the data is passed to the fuzzy system, it is multiplied by 0.01 to convert from kPa to bar.
    \item \textit{Fuzzy expert system.} The fuzzy expert system programming step is responsible for developing a server application to perform fuzzy control of the separator's pressure and to present the results in human-readable form. The control system is implemented in MATLAB \cite{fuzzy_matlab_link}, using the built-in Fuzzy Logic Toolbox to construct a Fuzzy Inference System (FIS), and SIMULINK to connect OPC communication and configure the FIS with its inputs and outputs. Additionally, Python is used with libraries such as NumPy \cite{numpy_citation} for mathematical operations on data arrays, SciKit-Fuzzy \cite{scikit_fuzzy_link_article} for creating FIS structures, Pandas \cite{pandas_link_citing} for convenient data storage in Excel files, and Matplotlib \cite{matplotlib_link_article} for plotting graphs.
    \item \textit{Metrics measurement.} Metrics measurement is used to evaluate how the system performs and to assess its accuracy with the applied control algorithm. It is the final crucial step in system verification, as it measures performance during experimental testing. The following metrics were applied: MSE, RMSE, MAE, IAE, ISE, and ITAE. In addition, to better assess the control behavior, we included the results for undershoot, overshoot, rise time, fall time, settling time, and steady-state error (SSE) for each case. Section~\ref{sec:evaluation_metrics} provides detailed information about each metric. Metrics were calculated for each initial pressure value, starting from 0 to 10 bar, with a step of 0.5 bar.
    \item \textit{User interface implementation.} The user interface implementation is the final step shown in the figure. This stage involves creating a web application that serves as an interface between the user and the control system. Streamlit is the primary library, as it enables fast, efficient development of front-end applications in Python~\cite{streamlit}. Figure~\ref{fig:streamlit-web-app} shows a screenshot of the homepage of the fuzzy split-range pressure control web application developed using Streamlit. The application is accessible via the web portal\footnote{https://temirbolat-fuzzy-split-range.streamlit.app}. The web interface allows users to configure simulation parameters, including: setpoint (target pressure in bar), initial pressure (starting pressure value in bar), total simulation time (duration in seconds), time step (interval between simulation steps), defuzzification method (Centroid, Bisector, MOM, LOM, or SOM), fuel valve gain (controls pressure increase), outlet valve gain (controls pressure drop), fuel flow (base inflow rate), base outflow (base outflow rate), noise (standard deviation of added noise to simulate disturbances), and an option to show or hide membership functions. Once the user completes the configuration and clicks the "Run simulation" button, two graphs are generated. The first graph shows how system pressure changes over time in response to control actions as the system attempts to reach the setpoint. The second graph shows how the positions of the fuel and outlet valves vary during the simulation to regulate the pressure. Below these graphs, a results table presents the performance metrics calculated using the selected defuzzification method. If the checkbox to display membership functions was selected, the corresponding plots are shown as well.
\end{itemize}

\section{Experimental Results}
\label{sec:experimental_results}

The evaluation was performed using two complementary approaches: real-time data transmission between the UniSim Design digital twin model and MATLAB Simulink via an OPC interface, and a standalone simulation implemented entirely in Python.

\subsection{Evaluation metrics}
\label{sec:evaluation_metrics}

It’s essential to evaluate how well the fuzzy control system works. To do this, we used several standard performance metrics. In our case, we focused on: Mean Squared Error (MSE), Root Mean Squared Error (RMSE), Mean Absolute Error (MAE), Integral of Absolute Error (IAE), Integral of Squared Error (ISE), and Integral of Time-weighted Absolute Error (ITAE). IAE, ISE, and ITAE are performance metrics commonly used in control systems \cite{IAE_ISE_ITAE_intro, IAE_ISE_ITAE_used}. Since we deal with control, we also measured overshoot or undershoot, rise or fall time, settling time, and steady-state error for each case. Table~\ref{table:evaluation_metrics_no_units_overview} stores main information about each mentioned metric.

\begin{table}[ht]
\caption{Information about applied evaluation metrics in the control system}
\label{table:evaluation_metrics_no_units_overview}
\begin{tabular*}{\textwidth}{@{\extracolsep\fill}p{3cm} p{5.9cm} c}
\toprule%
\centering Name & \centering Description & Formula \\
\midrule
Mean Squared Error (MSE) & Measures the average squared difference between predicted and actual values. Sensitive to large errors. & $\displaystyle \frac{1}{n} \sum_{i=1}^{n} \left( y_i - \hat{y}_i \right)^2$ \\
Root Mean Squared Error (RMSE) & Square root of MSE. Retains the original unit and penalizes large errors less harshly. & $\displaystyle \sqrt{ \frac{1}{n} \sum_{i=1}^{n} \left( y_i - \hat{y}_i \right)^2 }$ \\
Mean Absolute Error (MAE) & Average of absolute differences between predicted and actual values. Less sensitive to outliers. & $\displaystyle \frac{1}{n} \sum_{i=1}^{n} \left| y_i - \hat{y}_i \right|$ \\
Integral of Absolute Error (IAE) & Total accumulated error over time, regardless of direction. & $\displaystyle \int_0^{\infty} \left| e(t) \right| \, dt$  \\
Integral of Squared Error (ISE) & Accumulates squared error over time, emphasizing larger deviations. & $\displaystyle \int_0^{\infty} \left( e(t) \right)^2 \, dt$ \\
Integral of Time-weighted Absolute Error (ITAE) & Time-weighted error penalizing long-lasting deviations. & $\displaystyle \int_0^{\infty} t \cdot \left| e(t) \right| \, dt$ \\
Steady-State Error (SSE) & Final difference between the desired target and actual output value. & $\displaystyle \lim_{t \to \infty} \left| y(t) - y_{\text{target}} \right|$ \\
Rise Time (RT) & Time for system to rise from 10\% to 90\% of final value. & --- \\
Fall Time (FT) & Time for the system to fall from 90\% to 10\% of the final value. & --- \\
Settling Time (ST) & Time to reach and stay within 5\% (or 2\%) of the final value. & --- \\
Overshoot (O) & Maximum amount by which the output exceeds the target value. & $\displaystyle \frac{y_{\text{peak}} - y_{\text{target}}}{y_{\text{target}}} \times 100\%$ \\
Undershoot (U) & Maximum amount by which the output falls below the target value. & $\displaystyle \frac{y_{\text{target}} - y_{\text{min}}}{y_{\text{target}}} \times 100\%$ \\
\botrule
\end{tabular*}
\end{table}

Mean Squared Error (MSE) is a commonly used metric, especially when target values span a wide range. In control systems, the actual output values can differ noticeably from the desired ones due to system dynamics and external factors. This metric belongs to a group of regression metrics \cite{MSE}, which are used when the model predicts numeric values. In each formula from the Table~\ref{table:evaluation_metrics_no_units_overview}, $y_i$ is the expected output value at time step $i$, $\hat{y}_i$ is the actual output at time step $i$ and $n$ is the total number of records. MSE measures the average squared difference between the expected and actual values over the given interval. The best possible value of MSE is 0, indicating no error and perfect performance. In contrast, the worst value can grow infinitely large, indicating poor performance. It is a simple way to estimate the overall accuracy of the system; however, it does not provide any information about how quickly the system reaches the target value.

Root Mean Squared Error (RMSE) is another way to assess how accurately the system performs over time. It is based on MSE, but the main difference is that MSE gives more weight to significant errors and has a squared unit. RMSE takes the square root of MSE, so it’s smaller for substantial errors and has the same unit as the value we are measuring \cite{RMSE_MAE}.

Mean Absolute Error (MAE) is another regression metric that uses the L1 norm. It is less sensitive to significant errors, making it useful when the data contains outliers or noisy values \cite{MSE, RMSE_MAE}. Instead of calculating the squared difference, MAE measures the absolute difference between the predicted and actual values. It emphasizes the average deviation of the predictions and provides a stable, general measure of how a model behaves, even when a few bad data points occur during training.

The Integral of Absolute Error (IAE) is a performance metric that accumulates all errors over the entire time period, regardless of whether the error is positive or negative \cite{IAE_ISE_ITAE_theory}. Here, the term "error" refers to the difference between the desired value (setpoint) and the actual result (process value). This difference can be positive when the setpoint is higher than the process value, and negative when the setpoint is lower. The absolute value is used to ensure that the same errors, but with opposite signs, do not cancel each other, and the integral sums them all over time.

Integral of Squared Error (ISE) is a performance metric that sums the squares of the errors \cite{IAE_ISE_ITAE_theory}. Compared to other error metrics, ISE uses the square of the error. This means that when the system’s error at any moment is greater than 1, its contribution to the final ISE value increases rapidly \cite{ISE_theory}. As a result, ISE strongly penalizes significant errors and is highly sensitive to sudden changes in the system’s response.

The integral of Time-weighted Absolute Error (ITAE) is a performance metric similar to IAE. However, instead of simply summing absolute errors, it multiplies each absolute error by the time at which it occurs and then sums these values over the entire period \cite{IAE_ISE_ITAE_theory, ITAE_theory}. The key advantage of ITAE is that it punishes errors that persist over a long period of time. As time increases, if the error remains high or continues growing, the result of the time-error multiplication also increases. This makes ITAE a useful metric for evaluating how quickly the system responds to and corrects unexpected deviations or uncertainties.

Rise time is the time during which the system’s response increases from 10\% to 90\% of its final value \cite{rise_settling_time_theory}. In contrast, fall time measures how quickly the system’s output decreases from 90\% to 10\%. These measurements show how quickly the system’s output responds to the input over most of the time period. Settling time is the time it takes for the system’s response to reach and stay within a range of 5\% or 2\% of its final value \cite{rise_settling_time_theory}. The purpose of settling time is to demonstrate how long it takes the system to reach the acceptable range and stay within it.

Overshoot is the maximum value the system reaches during its transient response that exceeds the desired final value \cite{overshoot_theory}. According to the formula from the Table~\ref{table:evaluation_metrics_no_units_overview}, $y_{\text{peak}}$ is the maximum system's output and $y_{\text{target}}$ is the setpoint.

Undershoot is the lowest point at which the system’s response drops before reaching the desired target value. The term "overshoot" is used when the system’s output exceeds the target value. The purpose of measuring overshoot is to understand how much the system exceeds the desired target value. In contrast, "undershoot" shows how far the system’s output falls short of reaching the target. In the formula, $y_{\text{min}}$ is the minimum value the system came.

Steady-State Error (SSE) is the difference between the setpoint and the system’s final value \cite{rise_settling_time_theory}. It helps determine whether the system successfully reaches the desired output. The closer the SSE is to zero, the better the system performs in terms of control. In the formula, $y(t)$ is the system’s output at time $t$.

\subsection{Performance Evaluation}
% Тут про данные из Unisim Design

\begin{figure*}[htbp]
    \centering
    \begin{subfigure}[b]{0.483\textwidth}
        \includegraphics[width=\textwidth]{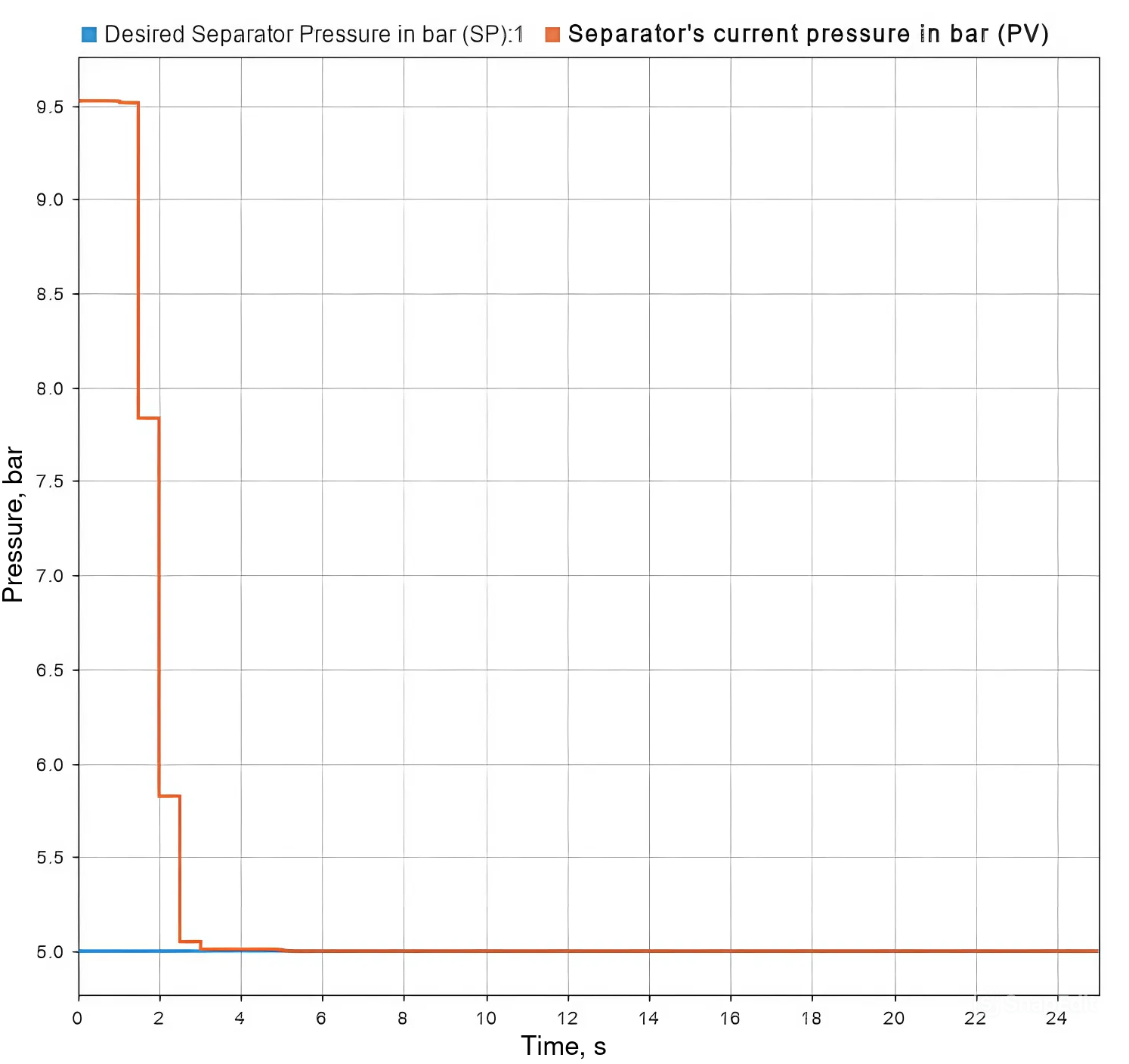}
        \caption{Bisector}
    \end{subfigure}
    \hfill
    \begin{subfigure}[b]{0.49\textwidth}
        \includegraphics[width=\textwidth]{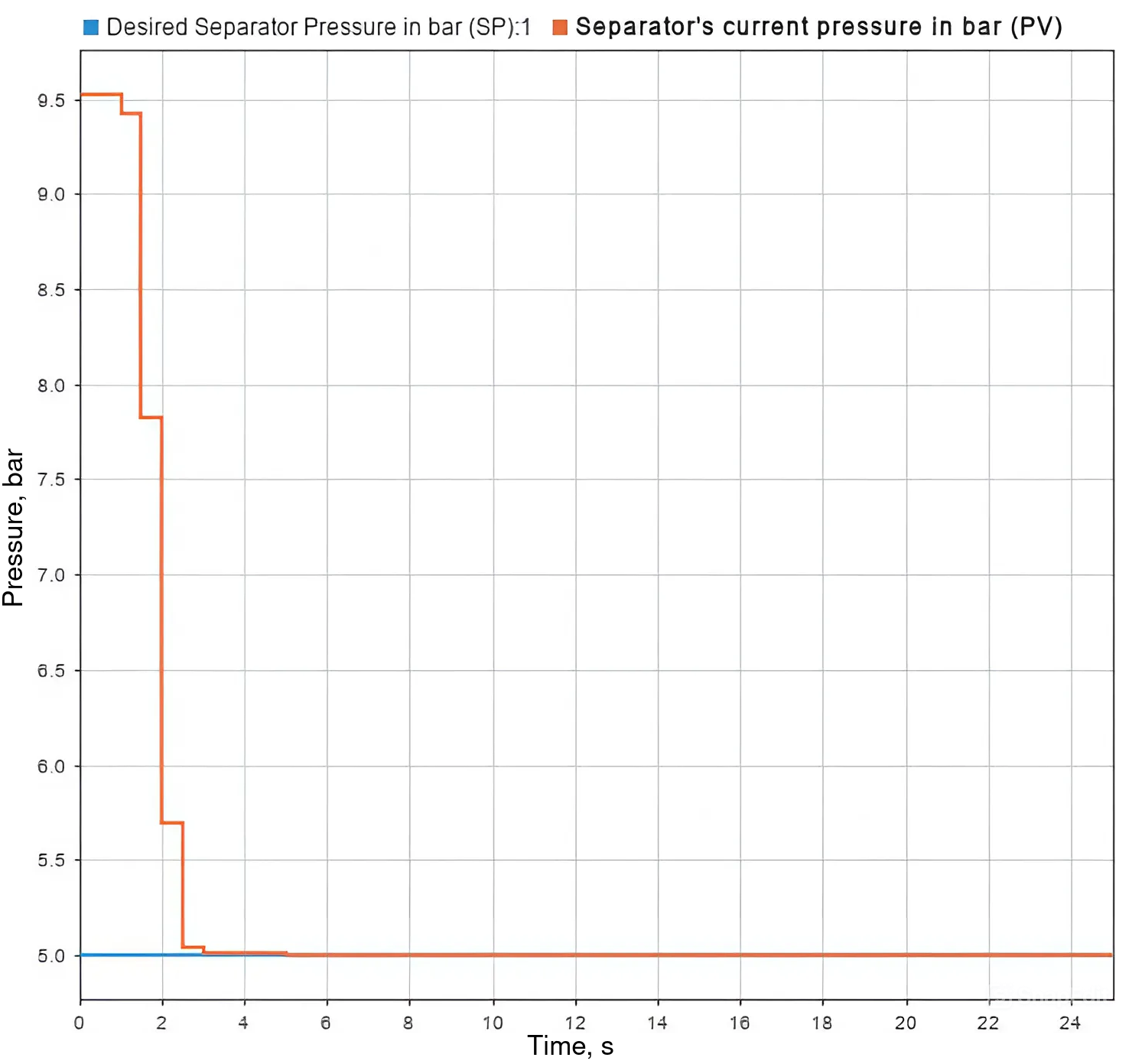}
        \caption{Centroid}
    \end{subfigure}
    \caption{Real-time fuzzy split-range pressure control with digital twin bisector and centroid defuzzification methods using MATLAB/SIMULINK starting from 9.53 bar}
    \label{fig:fuzzy_unisim_bisector_centroid_results}
\end{figure*}

\begin{figure*}[htbp]
    \centering
    \begin{subfigure}[b]{0.483\textwidth}
        \includegraphics[width=\textwidth]{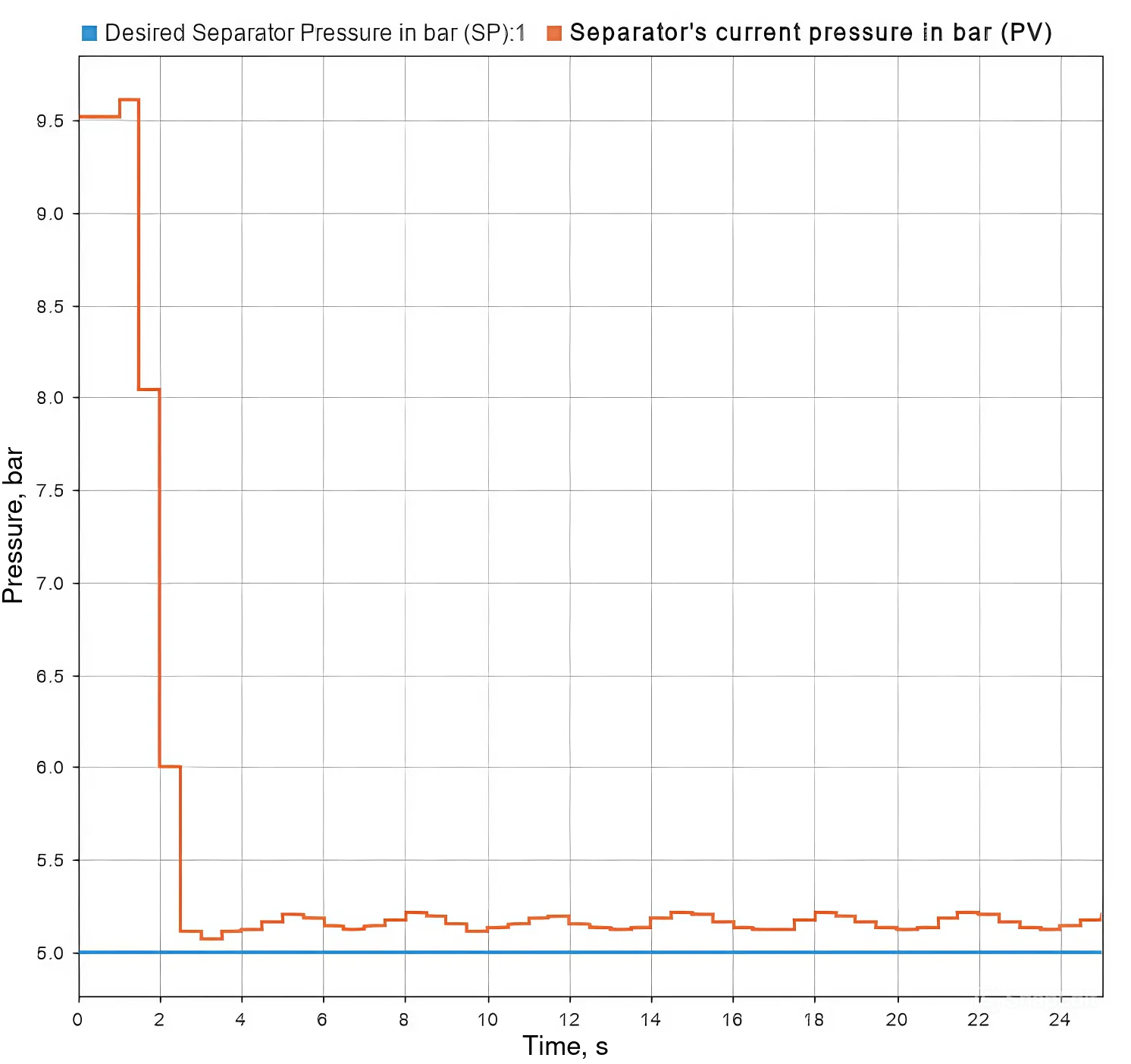}
        \caption{LOM}
    \end{subfigure}
    \hfill
    \begin{subfigure}[b]{0.49\textwidth}
        \includegraphics[width=\textwidth]{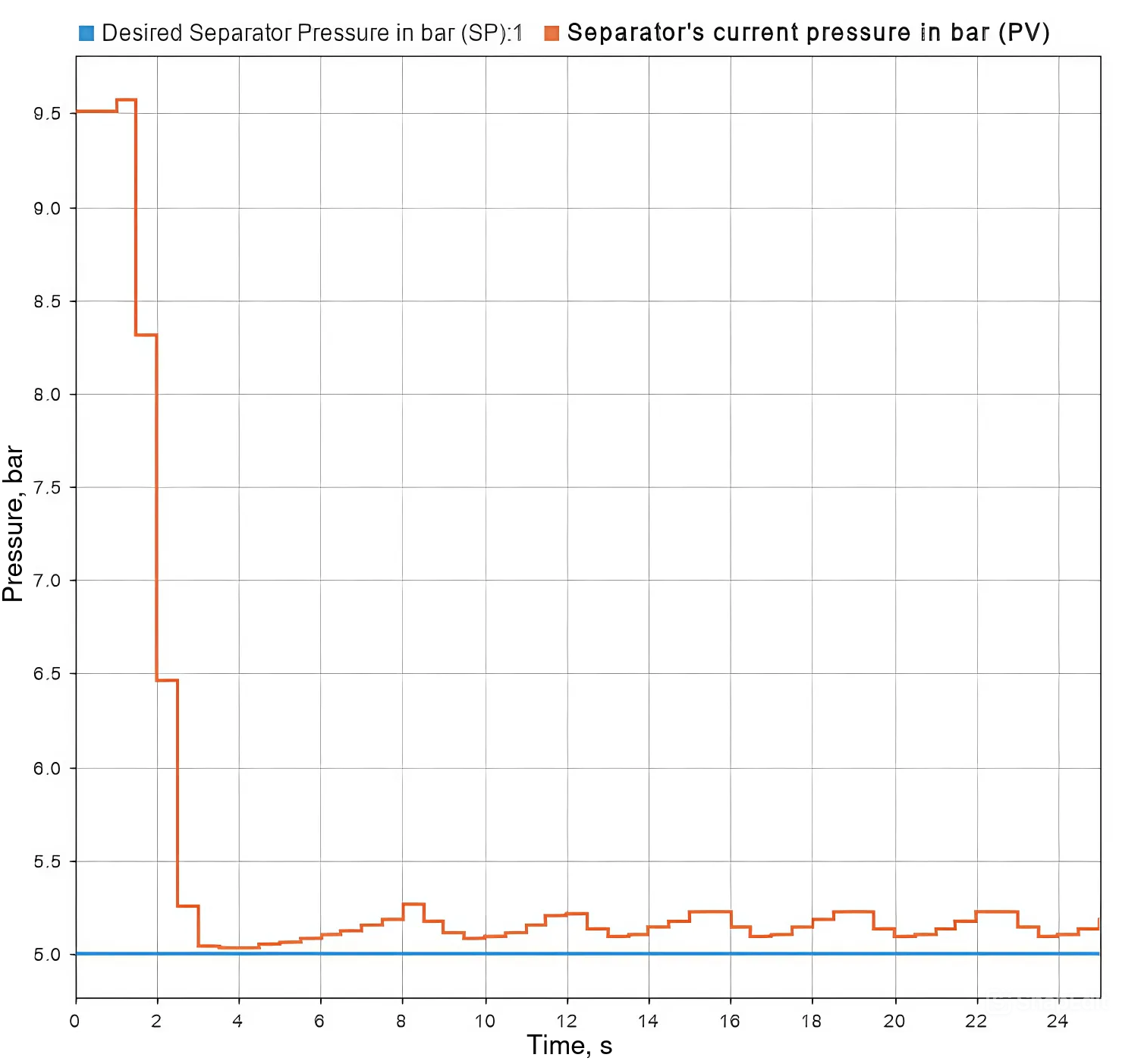}
        \caption{SOM}
    \end{subfigure}
    \hfill
    \begin{subfigure}[b]{0.49\textwidth}
        \includegraphics[width=\textwidth]{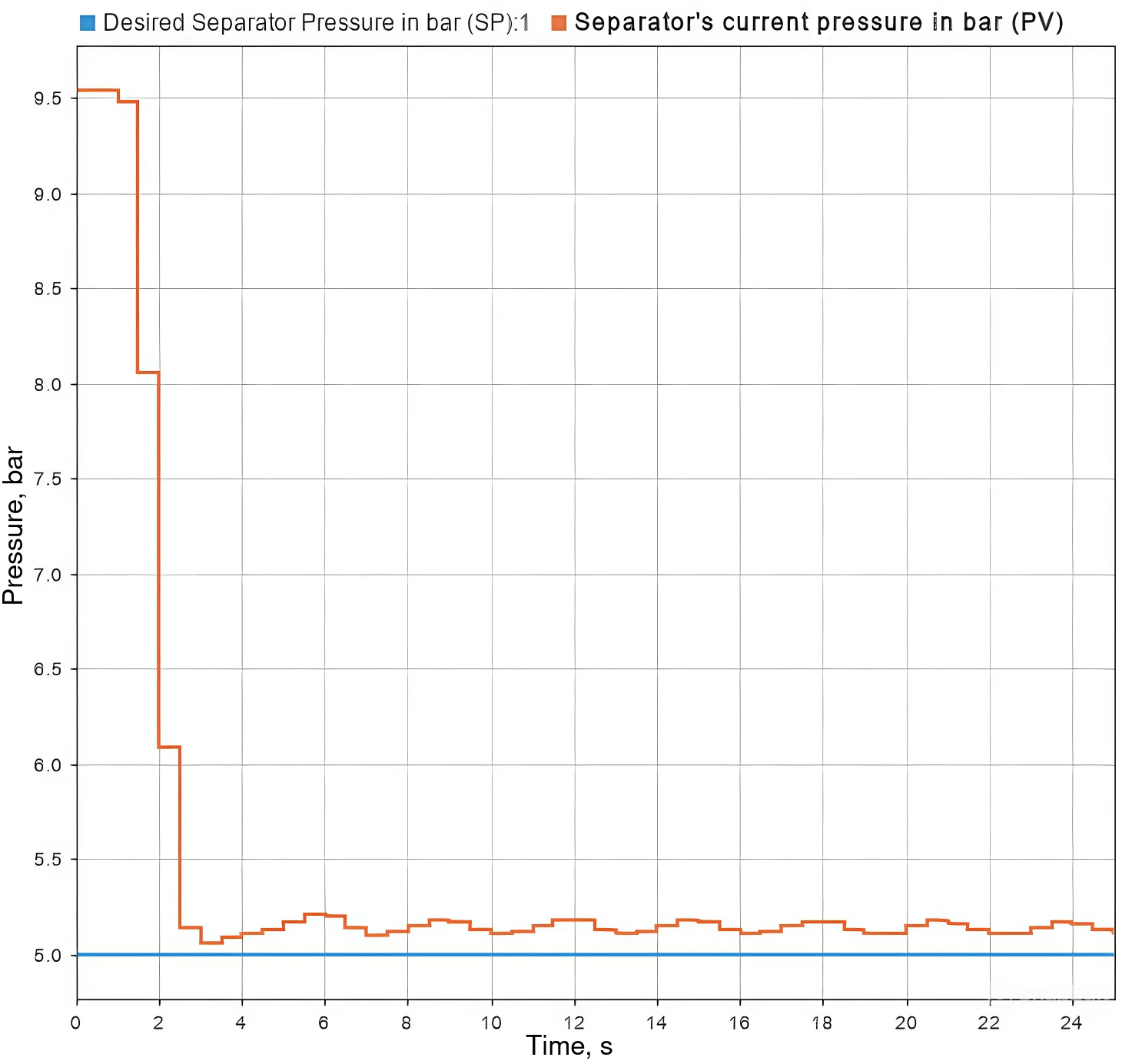}
        \caption{MOM}
    \end{subfigure}
    \caption{Real-time fuzzy split-range pressure control with digital twin under LOM, SOM, and MOM defuzzification methods using MATLAB/SIMULINK starting from 9.53 bar}
    \label{fig:fuzzy_unisim_lom_som_mom_results}
\end{figure*}

This section presents the results of real-time fuzzy pressure control experiments. Figure~\ref{fig:fuzzy_unisim_bisector_centroid_results} and Figure~\ref{fig:fuzzy_unisim_lom_som_mom_results} present five graphs illustrating the performance of the fuzzy controller over 25 seconds using different defuzzification methods: Largest of Maximum (LOM), Smallest of Maximum (SOM), Middle of Maximum (MOM), Bisector, and Centroid. Each graph includes two data lines: the separator’s desired pressure (blue) and the separator’s actual pressure (orange). The pressure setpoint is 5 bar for all cases, while the process values initially sit at approximately 9.5 bar. The actual pressure response appears more like a step due to the sampling time and a 0.5-second delay introduced by the valves. According to the figure, the graphs for the LOM, SOM, and MOM defuzzification methods fail to reach the desired value and exhibit continuous fluctuations near the setpoint, resulting in an unstable system. In contrast, the Bisector and Centroid methods nearly reach the target pressure range, with a deviation of about 1 \% or less from the setpoint, which is considered acceptable. However, the Centroid method exhibits a slightly smoother transition compared to the Bisector method.

% Тут описываем результаты с Питона с более идеальными условиями
To study the system’s behavior and evaluate the previously mentioned metrics for all rules and membership functions under various defuzzification methods, we developed a Python script and tested the results. The code checks how the rules respond to different error values ranging from -5 to 5, ensuring the coverage of boundary conditions for pressure control. It also enforces the molar flow balance, meaning that the amount entering the separator must equal the amount exiting. Additionally, slight random noise was added to the data using a normal (Gaussian) distribution with a mean of 0 and a standard deviation of 0.005. This approach introduces minor deviations around zero, simulating real-world measurement errors without significantly distorting the overall shape of the data. Table~\ref{table:evaluation_metrics_units} presents the units of measurement for each evaluation metric used to analyze performance in the context of pressure error measurement.

\begin{table}[ht]
    \caption{Units of Measurement for Evaluation Metrics}
    \label{table:evaluation_metrics_units}%
    \begin{tabular}{@{}lll@{}}
    \toprule
    Abbreviation & Metric name & Units \\
    \midrule
    MSE & Mean Squared Error & bar\textsuperscript{2} \\
    RMSE & Root Mean Squared Error & bar \\
    MAE & Mean Absolute Error & bar \\
    IAE & Integral of Absolute Error & bar$\cdot$s \\
    ISE & Integral of Squared Error & bar\textsuperscript{2}$\cdot$s \\
    ITAE & Integral of Time-weighted Absolute Error & bar$\cdot$s\textsuperscript{2} \\
    SSE & Steady-State Error & bar \\
    RT & Rise Time & s \\
    FT & Fall Time & s \\
    ST & Settling Time & s \\
    O/U & Overshoot/Undershoot & \% \\
    \botrule
    \end{tabular}
\end{table}

% \begin{table}[ht]
%     \centering
%     \caption{\textcolor{red}{Units of Measurement for Evaluation Metrics}} 
%     \begin{tabular}{|l|l|}
%     \hline 
%     \textbf{Metric name} & \textbf{Units} \\ 
%     \hline
%     MSE (Mean Squared Error) & bar\textsuperscript{2} \\ \hline
%     RMSE (Root Mean Squared Error) & bar \\ \hline
%     MAE (Mean Absolute Error) & bar \\ \hline
%     IAE (Integral of Absolute Error) & bar$\cdot$s \\ \hline
%     ISE (Integral of Squared Error) & bar\textsuperscript{2}$\cdot$s \\ \hline
%     ITAE (Integral of Time-weighted Absolute Error) & bar$\cdot$s\textsuperscript{2} \\ \hline
%     SSE (Steady-State Error) & bar \\ \hline
%     Rise Time & s \\ \hline
%     Fall Time & s \\ \hline
%     Settling Time & s \\ \hline
%     O/U (Overshoot/Undershoot) & \% \\ \hline
%     \end{tabular}
%     \label{table:evaluation_metrics_units}
% \end{table}

The evaluation metrics for each defuzzification method are divided into two categories: Error and Integral Metrics and Dynamic Metrics. The first group includes MSE, RMSE, MAE, IAE, ISE, and ITAE, which show how accurately the target pressure is achieved and how errors accumulate over time. The second group includes Steady-State Error (SSE), Rise Time, Fall Time, Settling Time, and Overshoot/Undershoot, which describe the system’s dynamic behavior and how quickly and consistently it responds to changes. The first column in each table represents the Initial Pressure Error (IPE), which is the difference between the setpoint (5 bar) and the process value, ranging from 0 to 10 bar. Therefore, the unit of PE is bar.

\begin{table}[ht]
\caption{LOM metrics results}
\label{table:lom_metrics_python}
\begin{tabular*}{\textwidth}{@{\extracolsep\fill}lccccccccccc}
\toprule%
& \multicolumn{5}{@{}c@{}}{Error and Integral Metrics} & \multicolumn{6}{@{}c@{}}{Dynamic metrics} \\ \cmidrule{1-7}\cmidrule{8-12}%
IPE & MSE & RMSE & MAE & IAE & ISE & ITAE & SSE & RT & FT & ST & O/U \\
\midrule
5.0 & 0.15 & 7.5 & - & - & 2.83 & 2.93 & 0.15 & 7.5 & - & - & 2.83 \\
4.5 & 2.16 & 1.47 & 0.82 & 20.52 & 54.09 & 86.16 & 0.14 & 6.9 & - & - & 2.30 \\
4.0 & 1.55 & 1.25 & 0.68 & 17.07 & 38.81 & 73.48 & 0.11 & 6.2 & - & - & 2.62 \\
3.5 & 1.07 & 1.04 & 0.56 & 14.01 & 26.87 & 63.28 & 0.14 & 5.6 & - & - & 2.96 \\
3.0 & 0.71 & 0.84 & 0.45 & 11.34 & 17.77 & 55.32 & 0.12 & 4.9 & - & - & 3.43 \\
2.5 & 0.44 & 0.66 & 0.36 & 8.99 & 10.90 & 49.43 & 0.13 & 4.2 & - & - & 4.12 \\
2.0 & 0.25 & 0.50 & 0.28 & 7.03 & 6.16 & 45.22 & 0.11 & 3.4 & - & - & 5.20 \\
1.5 & 0.12 & 0.34 & 0.21 & 5.31 & 2.90 & 42.22 & 0.11 & 2.5 & - & - & 6.97 \\
1.0 & 0.05 & 0.22 & 0.16 & 4.10 & 1.15 & 40.91 & 0.14 & 0.0 & - & - & 10.44 \\
0.5 & 0.02 & 0.14 & 0.14 & 3.40 & 0.51 & 40.41 & 0.12 & 0.0 & - & - & 20.63 \\
0.0 & 0.02 & 0.13 & 0.12 & 3.08 & 0.40 & 40.33 & 0.13 & - & 0.0 & - & 0.00 \\
-0.5 & 0.02 & 0.14 & 0.13 & 3.15 & 0.48 & 39.31 & 0.12 & - & 2.5 & - & 31.38 \\
-1.0 & 0.05 & 0.23 & 0.16 & 4.04 & 1.32 & 39.77 & 0.11 & - & 2.5 & - & 15.59 \\
-1.5 & 0.14 & 0.38 & 0.23 & 5.66 & 3.58 & 42.08 & 0.13 & - & 3.4 & - & 10.40 \\
-2.0 & 0.31 & 0.55 & 0.31 & 7.79 & 7.66 & 46.37 & 0.15 & - & 4.5 & - & 7.81 \\
-2.5 & 0.54 & 0.73 & 0.41 & 10.19 & 13.41 & 52.61 & 0.12 & - & 5.4 & - & 6.23 \\
-3.0 & 0.86 & 0.93 & 0.52 & 12.98 & 21.56 & 61.16 & 0.11 & - & 6.2 & - & 5.21 \\
-3.5 & 1.30 & 1.14 & 0.65 & 16.15 & 32.43 & 72.17 & 0.11 & - & 7.0 & - & 4.44 \\
-4.0 & 1.87 & 1.37 & 0.79 & 19.80 & 46.68 & 86.87 & 0.11 & - & 7.8 & - & 3.90 \\
-4.5 & 2.61 & 1.62 & 0.96 & 24.01 & 64.17 & 105.70 & 0.14 & - & 8.5 & - & 3.48 \\
-5.0 & 3.52 & 1.88 & 1.15 & 28.64 & 87.97 & 128.33 & 0.15 & - & 9.3 & - & 3.09 \\
\botrule
\end{tabular*}
\end{table}

Table~\ref{table:lom_metrics_python} shows the performance results for the LOM defuzzification method. The pressure control system shows major performance issues. According to the process documentation of the sour water treatment unit, the three-phase separator operates within a working pressure range of 4–6~bar, with about 5 bar being the nominal setpoint. For the representative case (IPE = 0), the RMSE value of 0.13 in normalized units corresponds to approximately $\pm0.26$ bar of deviation in the separator pressure within this operating range. This indicates that the fuzzy controller maintains pressure fluctuations within a physically realistic margin that complies with the plant’s design specifications. For positive initial pressure errors (IPE $>$ 0), the error metrics (MSE, RMSE, MAE) decrease gradually as the pressure approaches the setpoint. The best results are achieved near IPE = 0–0.5 bar, where MSE ranges from 0.016 to 0.020, RMSE from 0.126 to 0.143, and MAE from 0.123 to 0.136. For negative initial errors (IPE $<$ 0), the metrics increase symmetrically, with MSE reaching 3.519, IAE reaching 28.643, and ITAE reaching 128.329 at IPE = –5 bar. The ITAE metric becomes especially large at the edges because the initial error is high and remains high over time due to continuous fluctuations.

% \begin{table}[ht]
%     \centering
%     \caption{\textcolor{red}{LOM dynamic metrics results}} 
%     \begin{tabular}{|l|l|l|l|l|l|}
%     \hline \textbf{IPE} & \textbf{SSE} & \textbf{Rise time} & \textbf{Fall time}  & \textbf{Settling time} & \textbf{O/U} \\ \hline
%         5 & 0.153 & 7.5 & - & - & 2.830 \\ \hline
%         4.5 & 0.139 & 6.9 & - & - & 2.301 \\ \hline
%         4 & 0.111 & 6.2 & - & - & 2.622 \\ \hline
%         3.5 & 0.136 & 5.6 & - & - & 2.964 \\ \hline
%         3 & 0.122 & 4.9 & - & - & 3.431 \\ \hline
%         2.5 & 0.129 & 4.2 & - & - & 4.123 \\ \hline
%         2 & 0.112 & 3.4 & - & - & 5.204 \\ \hline
%         1.5& 0.113 & 2.5 & - & - & 6.974 \\ \hline
%         1 & 0.138 & 0.0 & - & - & 10.436 \\ \hline
%         0.5 & 0.115 & 0.0 & - & - & 20.625 \\ \hline
%         0 & 0.127 & - & 0.0 & - & 0.000 \\ \hline
%         -0.5 & 0.119 & - & 2.5 & - & 31.375 \\ \hline
%         -1 & 0.107 & - & 2.5 & - & 15.594 \\ \hline
%         -1.5 & 0.134 & - & 3.4 & - & 10.401 \\ \hline
%         -2 & 0.147 & - & 4.5 & - & 7.813 \\ \hline
%         -2.5 & 0.121 & - & 5.4 & - & 6.229 \\ \hline
%         -3 & 0.107 & - & 6.2 & - & 5.211 \\ \hline
%         -3.5 & 0.111 & - & 7.0 & - & 4.442 \\ \hline
%         -4 & 0.111 & - & 7.8 & - & 3.897 \\ \hline
%         -4.5 & 0.141 & - & 8.5 & - & 3.481 \\ \hline
%         -5 & 0.146 & - & 9.3 & - & 3.089 \\ \hline
%     \end{tabular}
%      \label{table:lom_dynamic_metrics_python}
% \end{table}

Overall, the system with the LOM method moves toward the setpoint. Still, when the error is positive (when pressure needs to increase), it never actually reaches the setpoint and keeps fluctuating near it. In cases of negative error (when pressure needs to decrease), the system crosses the setpoint, then stays slightly below it, and continues to fluctuate near it. SSE ranges between 0.107 and 0.153 bar. As a percentage relative to the setpoint, SSE is around 2.1-3.1\%, which is more than 2\% to get settling time.
Another critical point is that there is no settling time in any scenario, as the system oscillates around the setpoint without stabilizing. When the initial pressure error is small, the system shows a significant overshoot (up to 31.375 \%), indicating poor damping. However, for larger initial errors, the overshoot becomes much smaller, around 2.83\% and 3.09\%, respectively.

The performance evaluation results for the SOM defuzzification method are presented in Table~\ref{table:som_metrics_python}. As for the SOM error results, the fuzzy controller performs well for positive initial pressure errors (IPE $>$ 0): MSE, RMSE, and MAE decrease as the system approaches the setpoint. The most minor errors are observed around 0–0.5 bar, where MSE drops to 0.009, RMSE to 0.094, and MAE to 0.093. For negative initial errors (IPE $<$ 0), the errors start growing. At IPE = –5 bar, MSE rises to 3.583, and the integral metrics, such as IAE and ITAE, increase to 29.948 and 142.856, respectively, indicating that the system is more complex to remove significant excess pressure.

\begin{table}[ht]
\caption{SOM metrics results}
\label{table:som_metrics_python}
\begin{tabular*}{\textwidth}{@{\extracolsep\fill}lccccccccccc}
\toprule%
& \multicolumn{5}{@{}c@{}}{Error and Integral Metrics} & \multicolumn{6}{@{}c@{}}{Dynamic metrics} \\ \cmidrule{1-7}\cmidrule{8-12}%
IPE & MSE & RMSE & MAE & IAE & ISE & ITAE & SSE & RT & FT & ST & O/U \\
\midrule
5.0 & 2.93 & 1.71 & 0.96 & 24.10 & 73.18 & 95.35 & 0.10 & 7.6 & - & - & 1.93 \\
4.5 & 2.17 & 1.47 & 0.81 & 20.25 & 54.14 & 79.78 & 0.10 & 7.0 & - & - & 2.12 \\
4.0 & 1.56 & 1.25 & 0.67 & 16.79 & 38.87 & 66.96 & 0.10 & 6.4 & - & - & 2.48 \\
3.5 & 1.08 & 1.04 & 0.55 & 13.68 & 26.87 & 56.58 & 0.10 & 5.7 & - & - & 2.85 \\
3.0 & 0.72 & 0.85 & 0.44 & 11.01 & 17.89 & 47.96 & 0.10 & 5.0 & - & - & 3.27 \\
2.5 & 0.45 & 0.67 & 0.35 & 8.71 & 11.14 & 42.25 & 0.10 & 4.4 & - & - & 3.97 \\
2.0 & 0.26 & 0.51 & 0.27 & 6.71 & 6.42 & 37.43 & 0.10 & 3.6 & - & - & 4.89 \\
1.5 & 0.12 & 0.34 & 0.19 & 4.83 & 2.90 & 33.04 & 0.10 & 2.7 & - & - & 6.55 \\
1.0 & 0.04 & 0.20 & 0.14 & 3.48 & 1.02 & 31.73 & 0.10 & 11.4 & - & - & 9.45 \\
0.5 & 0.01 & 0.12 & 0.11 & 2.77 & 0.36 & 31.61 & 0.10 & 0.0 & - & - & 19.54 \\
0.0 & 0.01 & 0.09 & 0.09 & 2.31 & 0.22 & 30.85 & 0.10 & - & 0.0 & - & 0.00 \\
-0.5 & 0.02 & 0.15 & 0.15 & 3.62 & 0.58 & 42.49 & 0.13 & - & 0.2 & - & 24.55 \\
-1.0 & 0.06 & 0.24 & 0.18 & 4.58 & 1.48 & 43.32 & 0.14 & - & 0.4 & - & 12.20 \\
-1.5 & 0.17 & 0.41 & 0.26 & 6.44 & 4.14 & 46.59 & 0.13 & - & 3.7 & - & 8.17 \\
-2.0 & 0.36 & 0.60 & 0.36 & 9.01 & 9.03 & 53.06 & 0.13 & - & 5.1 & - & 6.11 \\
-2.5 & 0.60 & 0.77 & 0.46 & 11.44 & 14.88 & 60.83 & 0.13 & - & 6.1 & - & 4.91 \\
-3.0 & 0.93 & 0.96 & 0.57 & 14.27 & 23.13 & 70.69 & 0.12 & - & 6.9 & - & 4.08 \\
-3.5 & 1.36 & 1.17 & 0.70 & 17.45 & 33.99 & 83.14 & 0.14 & - & 7.7 & - & 3.50 \\
-4.0 & 1.93 & 1.39 & 0.85 & 21.11 & 48.28 & 99.03 & 0.14 & - & 8.6 & - & 3.04 \\
-4.5 & 2.66 & 1.63 & 1.01 & 25.26 & 66.57 & 118.65 & 0.13 & - & 9.3 & - & 2.74 \\
-5.0 & 3.58 & 1.89 & 1.20 & 29.95 & 89.56 & 142.86 & 0.13 & - & 10.1 & - & 2.45 \\
\botrule
\end{tabular*}
\end{table}

The SOM dynamic behavior shows that the rise time usually decreases as the pressure approaches the setpoint, except at IPE = 1 bar, where it suddenly jumps to 11.4 seconds, indicating instability.  The fall time increases as the pressure becomes more negative, reaching 10.1 seconds at IPE = –5 bar. The steady-state error stays relatively small in all cases, around 2–3\% of the 5 bar setpoint. The highest SSE is 0.144 at IPE = –3.5 bar, and the lowest is 0.098 at IPE = 0.5 bar.  Undershoot grows from about 2\% to 9\% for minor positive errors and reaches its peak of 19.5\% at IPE = 0.5 bar.  For negative errors, undershoot drops from 24.5\% to 2.4\% as the pressure decreases. In short, SOM works better than LOM, but it is still not a perfect solution. The system never fully reaches the setpoint, and there is no settling time, whether the error is positive or negative.

% \begin{table}[ht]
%     \centering
%     \caption{\textcolor{red}{SOM dynamic metrics results}} 
%     \begin{tabular}{|l|l|l|l|l|l|}
%     \hline \textbf{IPE} & \textbf{SSE} & \textbf{Rise time} & \textbf{Fall time}  & \textbf{Settling time} & \textbf{O/U} \\ \hline
%         5 & 0.100 & 7.6 & - & - & 1.925 \\ \hline
%         4.5 & 0.097 & 7.0 & - & - & 2.123 \\ \hline
%         4 & 0.100 & 6.4 & - & - & 2.484 \\ \hline
%         3.5 & 0.104 & 5.7 & - & - & 2.852 \\ \hline
%         3 & 0.101 & 5.0 & - & - & 3.267 \\ \hline
%         2.5 & 0.101 & 4.4 & - & - & 3.968 \\ \hline
%         2 & 0.099 & 3.6 & - & - & 4.893 \\ \hline
%         1.5 & 0.101 & 2.7 & - & - & 6.554 \\ \hline
%         1 & 0.103 & 11.4 & - & - & 9.447 \\ \hline
%         0.5 & 0.098 & 0.0 & - & - & 19.539 \\ \hline
%         0 & 0.100 & - & 0.0 & - & 0.00 \\ \hline
%         -0.5 & 0.126 & - & 0.2 & - & 24.553 \\ \hline
%         -1 & 0.142 & - & 0.4 & - & 12.203 \\ \hline
%         -1.5 & 0.125 & - & 3.7 & - & 8.171 \\ \hline
%         -2 & 0.133 & - & 5.1 & - & 6.114 \\ \hline
%         -2.5 & 0.126 & - & 6.1 & - & 4.910 \\ \hline
%         -3 & 0.124 & - & 6.9 & - & 4.080 \\ \hline
%         -3.5 & 0.144 & - & 7.7 & - & 3.501 \\ \hline
%         -4 & 0.141 & - & 8.6 & - & 3.042 \\ \hline
%         -4.5 & 0.125 & - & 9.3 & - & 2.735 \\ \hline
%         -5 & 0.128 & - & 10.1 & - & 2.446 \\ \hline
%     \end{tabular}
%      \label{table:som_dynamic_metrics_python}
% \end{table}

The results of the MOM error, integral, and dynamic metrics are shown in Table~\ref{table:mom_metrics_python}. For positive initial pressure errors, the controller shows behavior. The error metrics (MSE, RMSE, MAE) decrease as the system approaches the setpoint, but they remain too high near zero pressure error. The lowest errors are observed at IPE between 0 and 0.5 bar, with an MSE of 0.016, a RMSE of 0.125, and an MAE of 0.122. For negative initial errors (IPE $<$ 0), errors grow, especially after –2.5 bar. At IPE = –5 bar, MSE rises to 3.543, and the integral metrics IAE and ITAE reach 28.943 and 129.875, respectively, showing that significant excess pressures are harder to correct.

\begin{table}[ht]
\caption{MOM metrics results}
\label{table:mom_metrics_python}
\begin{tabular*}{\textwidth}{@{\extracolsep\fill}lccccccccccc}
\toprule%
& \multicolumn{5}{@{}c@{}}{Error and Integral Metrics} & \multicolumn{6}{@{}c@{}}{Dynamic metrics} \\ \cmidrule{1-7}\cmidrule{8-12}%
IPE & MSE & RMSE & MAE & IAE & ISE & ITAE & SSE & RT & FT & ST & O/U \\
\midrule
        5.0 & 2.92 & 1.71 & 0.98 & 24.45 & 73.04 & 102.82 & 0.15 & 7.5 & - & - & 2.13 \\
        4.5 & 2.17 & 1.47 & 0.83 & 20.62 & 54.17 & 87.00 & 0.12 & 6.9 & - & - & 2.36 \\
        4.0 & 1.56 & 1.25 & 0.69 & 17.15 & 38.90 & 73.94 & 0.14 & 6.3 & - & - & 2.65 \\
        3.5 & 1.08 & 1.04 & 0.56 & 14.12 & 26.95 & 64.07 & 0.13 & 5.7 & - & - & 3.03 \\
        3.0 & 0.72 & 0.85 & 0.46 & 11.43 & 17.88 & 55.90 & 0.14 & 4.9 & - & - & 3.54 \\
        2.5 & 0.44 & 0.66 & 0.36 & 9.07 & 11.01 & 49.97 & 0.15 & 4.2 & - & - & 4.23 \\
        2.0 & 0.26 & 0.51 & 0.29 & 7.16 & 6.36 & 45.70 & 0.13 & 3.5 & - & - & 5.32 \\
        1.5 & 0.12 & 0.34 & 0.22 & 5.40 & 2.96 & 43.02 & 0.15 & 2.6 & - & - & 7.08 \\
        1.0 & 0.05 & 0.22 & 0.16 & 4.11 & 1.17 & 40.94 & 0.11 & 0.3 & - & - & 10.55 \\
        0.5 & 0.02 & 0.14 & 0.14 & 3.43 & 0.52 & 40.72 & 0.11 & 0.2 & - & - & 21.22 \\
        0.0 & 0.02 & 0.13 & 0.12 & 3.05 & 0.39 & 40.44 & 0.15 & - & 0.0 & - & 0.000 \\
        -0.5 & 0.02 & 0.14 & 0.12 & 3.05 & 0.46 & 38.62 & 0.14 & - & 3.2 & - & 30.40 \\
        -1.0 & 0.05 & 0.23 & 0.16 & 3.98 & 1.33 & 38.51 & 0.12 & - & 3.9 & - & 15.16 \\
        -1.5 & 0.15 & 0.39 & 0.23 & 5.71 & 3.76 & 40.84 & 0.14 & - & 3.5 & - & 10.12 \\
        -2.0 & 0.33 & 0.57 & 0.32 & 8.06 & 8.24 & 46.14 & 0.12 & - & 4.7 & - & 7.60 \\
        -2.5 & 0.56 & 0.75 & 0.42 & 10.46 & 14.03 & 52.36 & 0.12 & - & 5.7 & - & 6.04 \\
        -3.0 & 0.89 & 0.94 & 0.53 & 13.28 & 22.23 & 61.66 & 0.13 & - & 6.5 & - & 5.05 \\
        -3.5 & 1.33 & 1.15 & 0.66 & 16.50 & 33.19 & 73.25 & 0.13 & - & 7.3 & - & 4.33 \\
        -4.0 & 1.90 & 1.38 & 0.81 & 20.14 & 47.44 & 88.07 & 0.15 & - & 8.1 & - & 3.80 \\ 
        -4.5 & 2.63 & 1.62 & 0.97 & 24.27 & 65.70 & 106.58 & 0.14 & - & 8.8 & - & 3.39 \\
        -5.0 & 3.54 & 1.88 & 1.16 & 28.94 & 88.59 & 129.88 & 0.15 & - & 9.6 & - & 3.02 \\
\botrule
\end{tabular*}
\end{table}

The MOM dynamic metrics show that rise time and fall time behave in a predictable manner.  Rise time decreases from 7.5 seconds at IPE = 5 bar to about 0.2 seconds around 0.5 bar. Fall time increases from 3.2 seconds to 9.6 seconds as IPE goes from –0.5 to –5 bar. There is no settling time, meaning the system does not settle within 2\% of the setpoint and continues fluctuating. The steady-state error (SSE) remains between 0.112 and 0.149 bar, corresponding to about 2–3\% across all cases. Overshoot rises from about 2\% to over 21\% for minor positive errors, with the highest value at IPE = 0.5 bar. For negative errors, undershoot starts high (30\% at –0.5 bar) and gradually drops to 3\% at –5 bar.
In conclusion, the MOM method shows decent results. It tends to overshoot the target pressure of 5 bar when starting from pressures higher than 5 bar, but then stays around that level with a stable, steady-state error. However, it exhibits large overshoots at minor positive errors and steady-state errors, leading to temporary pressure spikes.

% \begin{table}[ht]
%     \centering
%     \caption{\textcolor{red}{{MOM dynamic metrics results}}} 
%     \begin{tabular}{|l|l|l|l|l|l|}
%     \hline \textbf{IPE} & \textbf{SSE} & \textbf{Rise time} & \textbf{Fall time}  & \textbf{Settling time} & \textbf{O/U} \\ \hline
%         5 & 0.149 & 7.5 & - & - & 2.126 \\ \hline
%         4.5 & 0.122 & 6.9 & - & - & 2.364 \\ \hline
%         4 & 0.135 & 6.3 & - & - & 2.648 \\ \hline
%         3.5 & 0.129 & 5.7 & - & - & 3.027 \\ \hline
%         3 & 0.135 & 4.9 & - & - & 3.539 \\ \hline
%         2.5 & 0.146 & 4.2 & - & - & 4.228 \\ \hline
%         2 & 0.133 & 3.5 & - & - & 5.321 \\ \hline
%         1.5 & 0.152 & 2.6 & - & - & 7.082 \\ \hline
%         1 & 0.113 & 0.3 & - & - & 10.552 \\ \hline
%         0.5 & 0.112 & 0.2 & - & - & 21.215 \\ \hline
%         0 & 0.149 & - & 0.0 & - & 0.000 \\ \hline
%         -0.5 & 0.136 & - & 3.2 & - & 30.401 \\ \hline
%         -1 & 0.124 & - & 3.9 & - & 15.158 \\ \hline
%         -1.5 & 0.139 & - & 3.5 & - & 10.116 \\ \hline
%         -2 & 0.118 & - & 4.7 & - & 7.602 \\ \hline
%         -2.5 & 0.119 & - & 5.7 & - & 6.041 \\ \hline
%         -3 & 0.126 & - & 6.5 & - & 5.050 \\ \hline
%         -3.5 & 0.134 & - & 7.3 & - & 4.330 \\ \hline
%         -4 & 0.149 & - & 8.1 & - & 3.798 \\ \hline
%         -4.5 & 0.136 & - & 8.8 & - & 3.386 \\ \hline
%         -5 & 0.149 & - & 9.6 & - & 3.015 \\ \hline
%     \end{tabular}
%      \label{table:mom_dynamic_metrics_python}
% \end{table}

The performance results of the Bisector defuzzification method are shown in Table~\ref{table:bisector_metrics_python}. For positive initial pressure errors (IPE $>$ 0), MSE, RMSE, and MAE decrease as the pressure approaches the setpoint. The minimum errors occur around IPE = 0–0.5 bar, with MSE between 0.005 and 0.0, RMSE between 0.071 and 0.021, and MAE between 0.036 and 0.020. For negative initial errors, the errors increase smoothly and predictably. At IPE = -5 bar, MSE reaches 4.209, and integral errors such as IAE and ITAE rise to 32.898 and 150.196, respectively, indicating that the system behaves as expected, with the most significant errors occurring when the difference between the setpoint and the process value is greatest.

\begin{table}[ht]
\caption{Bisector metrics results}
\label{table:bisector_metrics_python}
\begin{tabular*}{\textwidth}{@{\extracolsep\fill}lccccccccccc}
\toprule%
& \multicolumn{5}{@{}c@{}}{Error and Integral Metrics} & \multicolumn{6}{@{}c@{}}{Dynamic metrics} \\ \cmidrule{1-7}\cmidrule{8-12}%
IPE & MSE & RMSE & MAE & IAE & ISE & ITAE & SSE & RT & FT & ST & O/U \\
\midrule
    5.0 & 3.35 & 1.83 & 1.04 & 25.98 & 83.84 & 94.03 & 0.02 & 8.5 & - & 10.5 & 0.39 \\ 
    4.5 & 2.47 & 1.57 & 0.85 & 21.36 & 61.66 & 71.74 & 0.02 & 7.8 & - & 9.6 & 0.40 \\
    4.0 & 1.77 & 1.33 & 0.69 & 17.30 & 44.20 & 53.97 & 0.02 & 7.0 & - & 8.6 & 0.47 \\
    3.5 & 1.22 & 1.10 & 0.55 & 13.73 & 30.48 & 40.38 & 0.02 & 6.3 & - & 7.7 & 0.56 \\
    3.0 & 0.80 & 0.89 & 0.42 & 10.50 & 19.95 & 28.68 & 0.02 & 5.4 & - & 6.7 & 0.62 \\
    2.5 & 0.48 & 0.70 & 0.31 & 7.71 & 12.07 & 20.28 & 0.02 & 4.6 & - & 5.7 & 0.73 \\
    2.0 & 0.26 & 0.51 & 0.21 & 5.23 & 6.39 & 13.76 & 0.02 & 3.7 & - & 4.6 & 0.89 \\
    1.5 & 0.11 & 0.33 & 0.13 & 3.22 & 2.75 & 9.95 & 0.02 & 2.8 & - & 3.4 & 1.31 \\
    1.0 & 0.03 & 0.18 & 0.07 & 1.75 & 0.84 & 7.42 & 0.02 & 1.9 & - & 2.2 & 1.64 \\
    0.5 & 0.01 & 0.07 & 0.04 & 0.90 & 0.13 & 7.05 & 0.02 & 1.4 & - & 1.1 & 3.84 \\
    0.0 & 0.00 & 0.02 & 0.02 & 0.50 & 0.011 & 6.60 & 0.02 & - & 0.0 & 0.0 & 0.00 \\
    -0.5 & 0.01 & 0.08 & 0.04 & 0.98 & 0.17 & 6.83 & 0.02 & - & 2.4 & 1.7 & 4.92 \\
    -1.0 & 0.05 & 0.21 & 0.09 & 2.16 & 1.14 & 8.22 & 0.02 & - & 3.0 & 3.4 & 2.36 \\
    -1.5 & 0.15 & 0.39 & 0.17 & 4.13 & 3.70 & 12.17 & 0.02 & - & 3.9 & 4.9 & 1.53 \\
    -2.0 & 0.35 & 0.59 & 0.27 & 6.85 & 8.62 & 19.49 & 0.02 & - & 5.1 & 6.5 & 1.12 \\
    -2.5 & 0.64 & 0.80 & 0.40 & 10.10 & 16.06 & 30.86 & 0.02 & - & 6.2 & 7.9 & 0.94 \\
    -3.0 & 1.04 & 1.02 & 0.55 & 13.69 & 26.07 & 45.65 & 0.02 & - & 7.4 & 9.2 & 0.77 \\
    -3.5 & 1.57 & 1.25 & 0.71 & 17.68 & 39.18 & 63.99 & 0.02 & - & 8.4 & 10.5 & 0.65 \\
    -4.0 & 2.24 & 1.50 & 0.88 & 22.08 & 56.97 & 86.23 & 0.02 & - & 9.3 & 11.6 & 0.57 \\
    -4.5 & 3.11 & 1.77 & 1.09 & 27.17 & 77.76 & 114.96 & 0.02 & - & 10.2 & 12.8 & 0.51 \\
    -5.0 & 4.21 & 2.05 & 1.32 & 32.9 & 105.22 & 150.20 & 0.02 & - & 11.2 & 14.0 & 0.44 \\
\botrule
\end{tabular*}
\end{table}

The Bisector dynamic metrics show that rise time decreases as the error approaches zero, dropping from 8.5 seconds (at IPE = 5 bar) to 1.4 seconds (at IPE = 0.5 bar). Fall time steadily increases from 2.4 seconds (at IPE = –0.5 bar) to 11.2 seconds (at IPE = –5 bar). The steady-state error (SSE) remains low across all cases, ranging from 0.019 to 0.023 bar. When the process value is close to the 5-bar setpoint, the SSE is around 0.4–0.5\%, indicating excellent steady-state performance. Compared to LOM, SOM, and MOM, the Bisector method has much smaller overshoot and undershoot. For positive IPE, the overshoot stays below 4\%, peaking at 3.84\% when IPE = 0.5 bar. For negative IPE, the undershoot drops smoothly from 4.9\% to 0.4\% as IPE moves from –0.5 to –5 bar. Settling time is present in all cases and stays reasonable, meaning the system stabilizes within 2\% of the setpoint. For minor initial errors, stabilization occurs within 1–3 seconds, whereas for larger errors, the settling time increases but remains acceptable, reaching 14 seconds in the worst case (at IPE = –5 bar). In summary, the Bisector method shows excellent final accuracy, minimal overshoot, low steady-state errors, and smooth dynamic behavior even with large initial deviations.

% \begin{table}[ht]
%     \centering
%     \caption{\textcolor{red}{Bisector dynamic metrics results}} 
%     \begin{tabular}{|l|l|l|l|l|l|}
%     \hline \textbf{IPE} & \textbf{SSE} & \textbf{Rise time} & \textbf{Fall time}  & \textbf{Settling time} & \textbf{O/U} \\ \hline
%         5 & 0.021 & 8.5 & - & 10.5 & 0.385 \\ \hline
%         4.5 & 0.023 & 7.8 & - & 9.6 & 0.395 \\ \hline
%         4 & 0.019 & 7.0 & - & 8.6 & 0.468 \\ \hline
%         3.5 & 0.023 & 6.3 & - & 7.7 & 0.557 \\ \hline
%         3 & 0.020 & 5.4 & - & 6.7 & 0.615 \\ \hline
%         2.5 & 0.019 & 4.6 & - & 5.7 & 0.732 \\ \hline
%         2 & 0.020 & 3.7 & - & 4.6 & 0.888 \\ \hline
%         1.5 & 0.022 & 2.8 & - & 3.4 & 1.314 \\ \hline
%         1 & 0.021 & 1.9 & - & 2.2 & 1.636 \\ \hline
%         0.5 & 0.022 & 1.4 & - & 1.1 & 3.84 \\ \hline
%         0 & 0.019 & - & 0.0 & 0.0 & 0.00 \\ \hline
%         -0.5 & 0.023 & - & 2.4 & 1.7 & 4.916 \\ \hline
%         -1 & 0.021 & - & 3.0 & 3.4 & 2.362 \\ \hline
%         -1.5 & 0.022 & - & 3.9 & 4.9 & 1.526 \\ \hline
%         -2 & 0.020 & - & 5.1 & 6.5 & 1.116 \\ \hline
%         -2.5 & 0.022 & - & 6.2 & 7.9 & 0.938 \\ \hline
%         -3 & 0.022 & - & 7.4 & 9.2 & 0.774  \\ \hline
%         -3.5 & 0.022 & - & 8.4 & 10.5 & 0.647 \\ \hline
%         -4 & 0.022 & - & 9.3 & 11.6 & 0.571 \\ \hline
%         -4.5 & 0.022 & - & 10.2 & 12.8 & 0.509 \\ \hline
%         -5 & 0.021 & - & 11.2 & 14.0 & 0.438 \\ \hline
%     \end{tabular}
%      \label{table:bisector_dynamic_metrics_python}
% \end{table}

The performance results of the Centroid defuzzification method are shown in Table~\ref{table:centroid_metrics_python}. According to the error metric results, for positive initial pressure errors (IPE $>$ 0), MSE, RMSE, and MAE decrease smoothly as the system approaches the setpoint. The most minor errors occur near IPE = 0–0.5 bar, where MSE drops to 0.005–0.0, RMSE to 0.069–0.001, and MAE to 0.016–0.001. For negative initial errors (IPE $<$ 0), errors and metric values increase. Even at IPE = –5 bar, MSE reaches 4.342, IAE is 33.771, and ITAE is 156.285 — which are acceptable values considering the large initial deviations. Compared to the Bisector method, the Centroid method shows a slightly higher ITAE because it reduces the error more gradually, step by step, which takes longer.

\begin{table}[ht]
\caption{Centroid metrics results}
\label{table:centroid_metrics_python}
\begin{tabular*}{\textwidth}{@{\extracolsep\fill}lccccccccccc}
\toprule%
& \multicolumn{5}{@{}c@{}}{Error and Integral Metrics} & \multicolumn{6}{@{}c@{}}{Dynamic metrics} \\ \cmidrule{1-7}\cmidrule{8-12}%
IPE & MSE & RMSE & MAE & IAE & ISE & ITAE & SSE & RT & FT & ST & O/U \\
\midrule
    5.0 & 3.44 & 1.86 & 1.05 & 26.35 & 86.00 & 93.33 & 0.001 & 8.7 & - & 10.9 & 0.02  \\
    4.5 & 2.54 & 1.59 & 0.87 & 21.65 & 63.55 & 70.02 & 0.000 & 8 & - & 9.9 & 0.01 \\
    4.0 & 1.82 & 1.35 & 0.70 & 17.40 & 45.43 & 50.85 & 0.000 & 7.2 & - & 8.9 & 0.03 \\
    3.5 & 1.26 & 1.12 & 0.55 & 13.72 & 31.48 & 35.96 & 0.001 & 6.3 & - & 7.9 & 0.02 \\
    3.0 & 0.83 & 0.91 & 0.42 & 10.43 & 20.62 & 23.08 & 0.000 & 5.6 & - & 6.9 & 0.01 \\
    2.5 & 0.49 & 0.70 & 0.30 & 7.41 & 12.27 & 14.51 & 0.000 & 4.7 & - & 5.9 & 0.05 \\
    2.0 & 0.26 & 0.51 & 0.19 & 4.83 & 6.41 & 7.78 & 0.001 & 3.8 & - & 4.7 & 0.08 \\
    1.5 & 0.11 & 0.33 & 0.11 & 2.78 & 2.74 & 3.60 & 0.000 & 2.9 & - & 3.5 & 0.03 \\
    1.0 & 0.03 & 0.18 & 0.05 & 1.31 & 0.84 & 1.36 & 0.001 & 2.1 & - & 2.4 & 0.11 \\
    0.5 & 0.01 & 0.07 & 0.02 & 0.40 & 0.12 & 0.52 & 0.000 & 1.3 & - & 1.2 & 0.30 \\
    0.0 & 0.00 & 0.00 & 0.00 & 0.02 & 0.00 & 0.29 & 0.001 & - & 0.0 & 0.0 & 0.00 \\ 
    -0.5 & 0.01 & 0.08 & 0.02 & 0.55 & 0.16 & 0.71 & 0.000 & - & 2.0 & 1.7 & 0.56 \\
    -1.0 & 0.05 & 0.21 & 0.07 & 1.79 & 1.14 & 2.39 & 0.002 & - & 3.0 & 3.4 & 0.27 \\ 
    -1.5 & 0.15 & 0.39 & 0.15 & 3.80 & 3.73 & 6.58 & 0.000 & - & 4.1 & 4.9 & 0.18 \\
    -2.0 & 0.35 & 0.59 & 0.26 & 6.57 & 8.64 & 14.39 & 0.002 & - & 5.2 & 6.5 & 0.18 \\
    -2.5 & 0.66 & 0.81 & 0.40 & 10.01 & 16.43 & 26.78 & 0.001 & - & 6.4 & 8.1 & 0.12 \\
    -3.0 & 1.09 & 1.05 & 0.56 & 13.92 & 27.29 & 43.42 & 0.001 & - & 7.5 & 9.5 & 0.08 \\
    -3.5 & 1.64 & 1.28 & 0.72 & 18.10 & 40.99 & 63.50 & 0.002 & - & 8.6 & 10.7 & 0.09 \\
    -4.0 & 2.34 & 1.53 & 0.91 & 22.72 & 58.39 & 88.34 & 0.001 & - & 9.6 & 12.0 & 0.07 \\
    -4.5 & 3.23 & 1.80 & 1.12 & 27.93 & 80.75 & 118.77 & 0.001 & - & 10.5 & 13.2 & 0.07 \\
    -5.0 & 4.34 & 2.08 & 1.35 & 33.77 & 108.55 & 156.29 & 0.001 & - & 11.4 & 14.4 & 0.05 \\
\botrule
\end{tabular*}
\end{table}

The Centroid dynamic metrics show that rise time decreases from 8.7 seconds (at IPE = 5) to 1.3 seconds (at IPE = 0.5), while fall time increases from 2.0 seconds (at IPE = –0.5) to 11.4 seconds (at IPE = –5). This behavior is typical for significant pressure corrections. The SSE remains extremely low across all cases and among all defuzzification methods. It ranges between 0.0 and 0.002 bar, corresponding to less than 0.05\% of the 5 bar setpoint. Overshoot and undershoot are also well within acceptable ranges. For positive IPE, overshoot is very low, reaching only 0.297\% at IPE = 0.5 bar. For negative IPE, the undershoot remains low, reaching a maximum of 0.555\% at IPE = –0.5 bar, and decreases further with larger initial errors. Settling time behaves as follows: minor initial errors are corrected within 1–3 seconds, while large initial deviations take about 14 seconds in the worst case (at IPE = –5). In conclusion, the Centroid method demonstrates the best steady-state accuracy, minimal dynamic overshoot, fast settling times, and smooth behavior across the entire range of initial errors.

% \begin{table}[ht]
%     \centering
%     \caption{\textcolor{red}{Centroid dynamic metrics results}} 
%     \begin{tabular}{|l|l|l|l|l|l|}
%     \hline \textbf{IPE} & \textbf{SSE} & \textbf{Rise time} & \textbf{Fall time}  & \textbf{Settling time} & \textbf{O/U} \\ \hline
%         5 & 0.001 & 8.7 & - & 10.9 & 0.021 \\ \hline
%         4.5 & 0.000 & 8 & - & 9.9 & 0.014 \\ \hline
%         4 & 0.000 & 7.2 & - & 8.9 & 0.026 \\ \hline
%         3.5 & 0.001 & 6.3 & - & 7.9 & 0.020 \\ \hline
%         3 & 0.000 & 5.6 & - & 6.9 & 0.014 \\ \hline
%         2.5 & 0.000 & 4.7 & - & 5.9 & 0.048 \\ \hline
%         2 & 0.001 & 3.8 & - & 4.7 & 0.080 \\ \hline
%         1.5 & 0.000 & 2.9 & - & 3.5 & 0.030 \\ \hline
%         1 & 0.001 & 2.1 & - & 2.4 & 0.113 \\ \hline
%         0.5 & 0.000 & 1.3 & - & 1.2 & 0.297 \\ \hline
%         0 & 0.001 & - & 0.0 & 0.0 & 0.000 \\ \hline
%         -0.5 & 0.000 & - & 2.0 & 1.7 & 0.555 \\ \hline
%         -1 & 0.002 & - & 3.0 & 3.4 & 0.266 \\ \hline
%         -1.5 & 0.000 & - & 4.1 & 4.9 & 0.180 \\ \hline
%         -2 & 0.002 & - & 5.2 & 6.5 & 0.180 \\ \hline
%         -2.5 & 0.001 & - & 6.4 & 8.1 & 0.123 \\ \hline
%         -3 & 0.001 & - & 7.5 & 9.5 & 0.084 \\ \hline
%         -3.5 & 0.002 & - & 8.6 & 10.7 & 0.091 \\ \hline
%         -4 & 0.001 & - & 9.6 & 12.0 & 0.066 \\ \hline
%         -4.5 & 0.001 & - & 10.5 & 13.2 & 0.072 \\ \hline
%         -5 & 0.001 & - & 11.4 & 14.4 & 0.048 \\ \hline
%     \end{tabular}
%      \label{table:centroid_dynamic_metrics_python}
% \end{table}

\begin{figure*}[htbp]
    \centering
    \begin{subfigure}[b]{0.49\textwidth}
        \includegraphics[width=\textwidth]{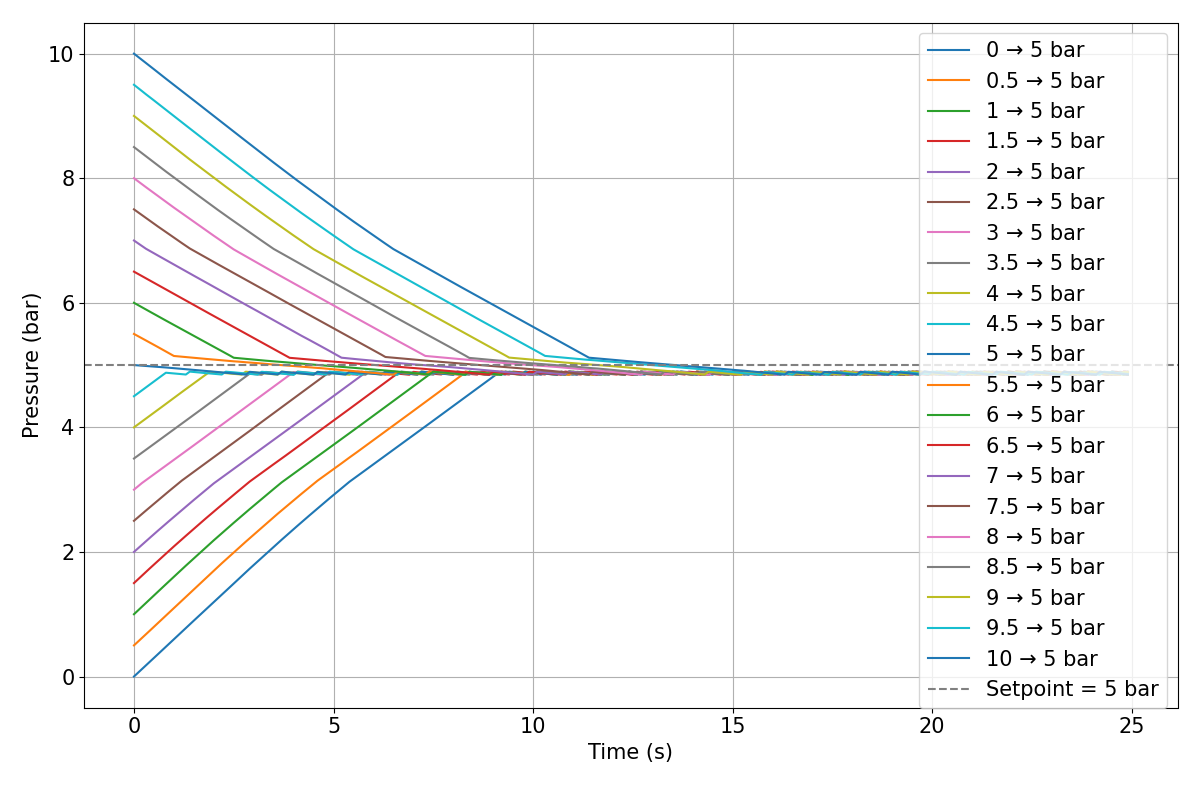}
        \caption{LOM}
    \end{subfigure}
    \hfill
    \begin{subfigure}[b]{0.49\textwidth}
        \includegraphics[width=\textwidth]{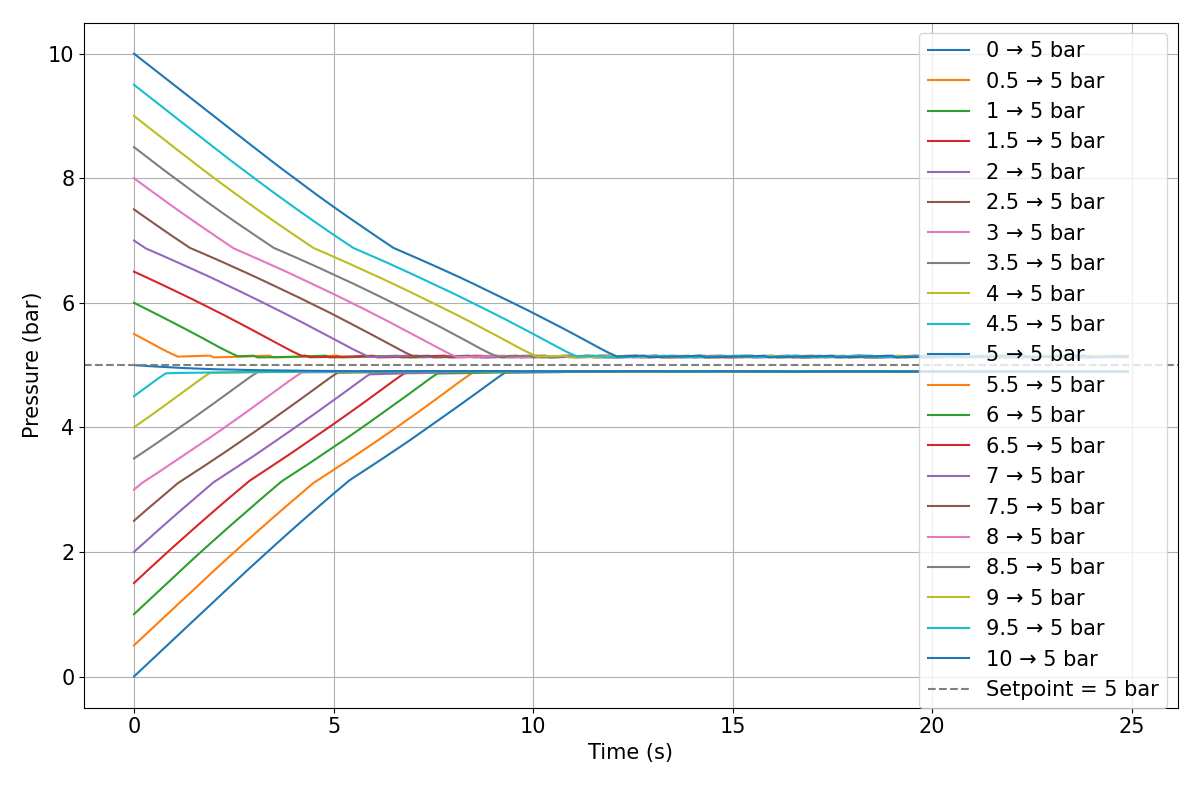}
        \caption{SOM}
    \end{subfigure}
    \hfill
    \begin{subfigure}[b]{0.49\textwidth}
        \includegraphics[width=\textwidth]{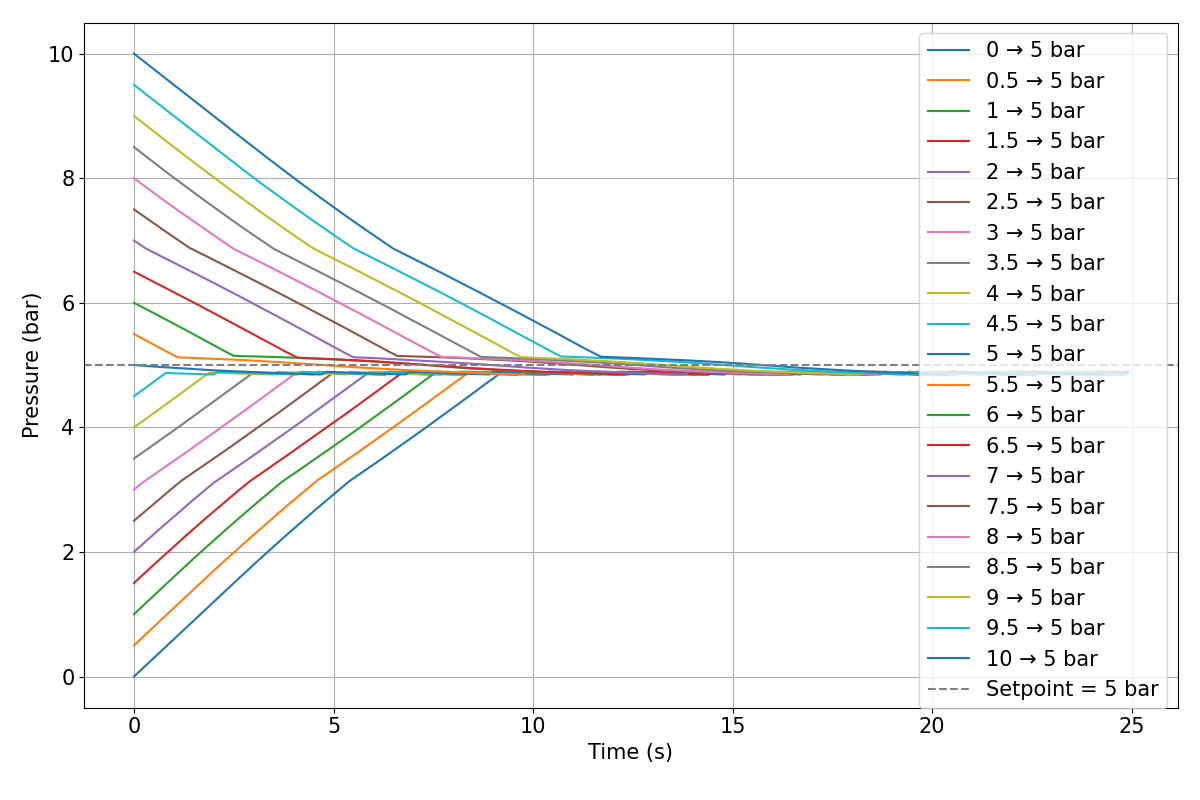}
        \caption{MOM}
    \end{subfigure}
    \hfill
    \begin{subfigure}[b]{0.49\textwidth}
        \includegraphics[width=\textwidth]{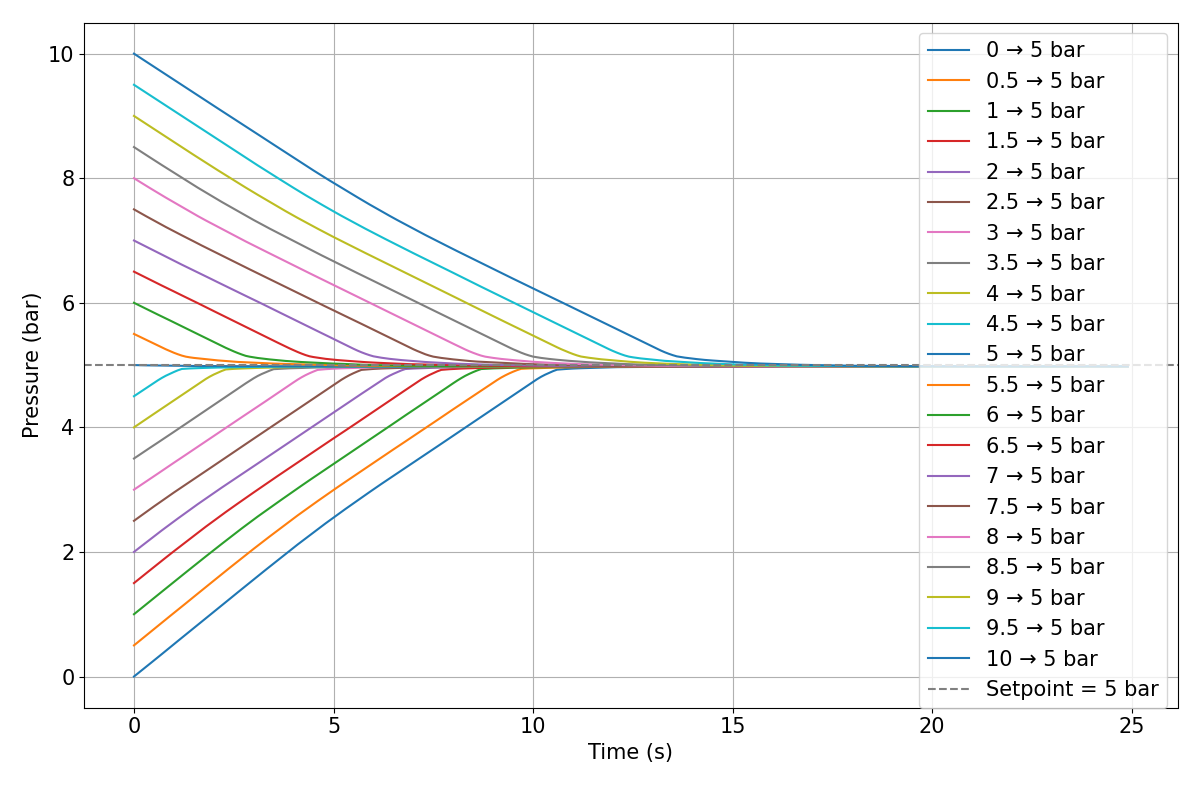}
        \caption{Bisector}
    \end{subfigure}
    \hfill
    \begin{subfigure}[b]{0.5\textwidth}
        \includegraphics[width=\textwidth]{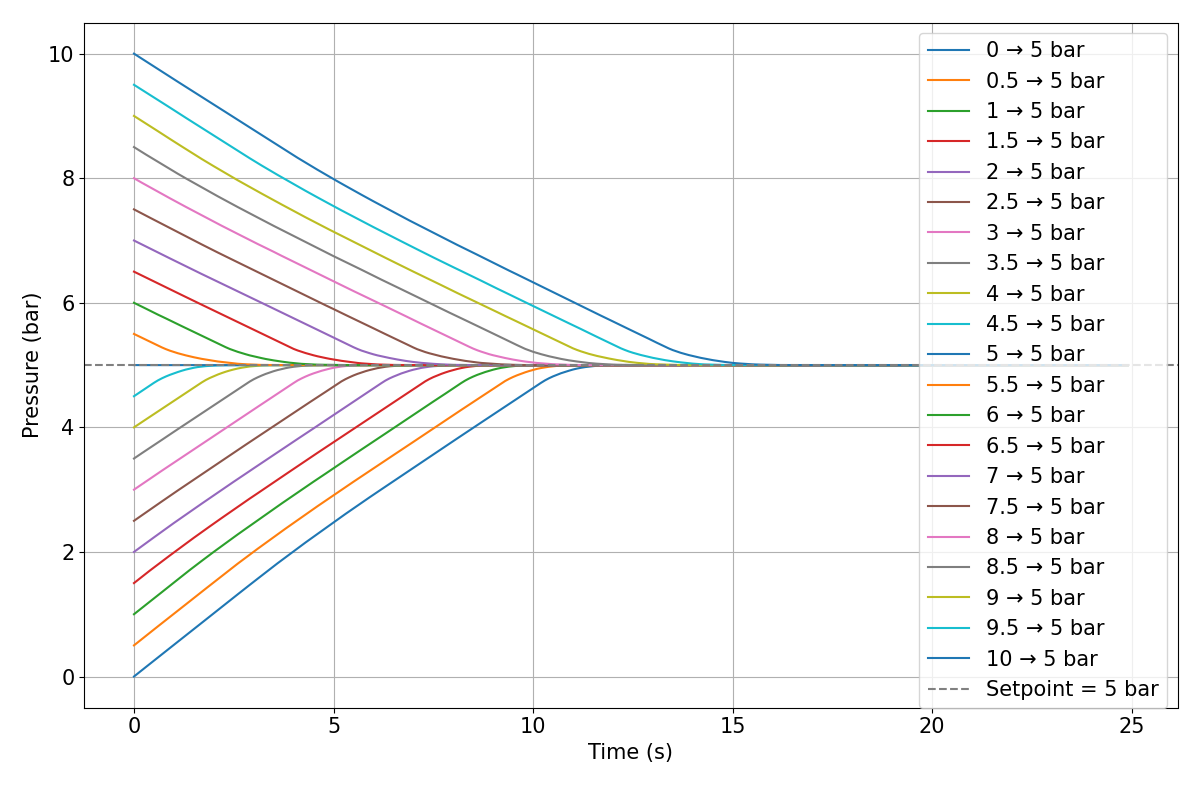}
        \caption{Centroid}
    \end{subfigure}
    \caption{Python simulation of fuzzy split-range pressure control with various methods}
    \label{fig:fuzzy_split_range_python_combined}
\end{figure*}

Overall, each defuzzification method has its own strengths and weaknesses. Figure~\ref{fig:fuzzy_split_range_python_combined} shows the visual results for each defuzzification method under different initial pressure values, ranging from 0 to 10 bar in 0.5 bar steps. The Bisector and Centroid methods achieved the best pressure control results. Centroid provides the best final accuracy, with minimal steady-state error and overshoot, while the Bisector achieves stability faster with a shorter settling time. The LOM, SOM, and MOM methods try to reach the setpoint more quickly, but none of them manage to settle within 2\% of the setpoint. As a result, they continue fluctuating and never stabilize, leading to significant steady-state errors. One possible approach is to combine these methods: initially using SOM, MOM, or LOM to reduce substantial errors quickly, and then switching to the Centroid or Bisector methods for a smoother response and precise final stabilization.

\section{Conclusion}
\label{sec:conclusion}

% Intro paraphrasing
The process of collecting and purifying sour water is crucial in the oil and gas sector, as it reduces acid and carbon dioxide emissions into the environment and increases production efficiency by promoting the rational use of natural resources through the additional purification and reuse of harmful components.

% Main message
This paper presents a new expert system based on fuzzy logic for optimal control of key parameters, using five defuzzification methods to compare their performance and identify the most effective one. Pressure split-range control in a three-phase separator is used as an example. To enhance accessibility and usability, an interactive web-based simulation tool was developed using the Streamlit framework. This tool allows users to adjust parameters, run simulations, visualize results, and evaluate performance metrics of the fuzzy control system.

The best results were achieved with the Centroid and Bisector defuzzification methods, with the Centroid method providing slightly higher accuracy. A custom digital twin of the technological process, built using Honeywell's Unisim Design, accurately replicates the physical system, enabling efficient data collection and analysis in the oil and gas sector. The control approach allows junior staff without deep technical knowledge to operate the system effectively, as it mimics human reasoning based on documentation describing the facility's technological process.

The proposed system contributes to industrial safety by minimizing direct human intervention in potentially hazardous acidic water treatment processes, thereby reducing the risk of accidents and exposure.

% Limitation
The system has its limitations. Currently, it has sufficient control rules, but to achieve smoother transitions, it would be better to extend the system by adding more membership functions at both the input and output, along with additional rules. Moreover, the current experiment applies fuzzy logic to maintain the parameters at desired values, but further work is needed to compare this control method with other approaches, such as PID regulation. Additional experiments should also be conducted to ensure there are no unexpected results, as the current testing covers only initial pressures from 0 to 10 bar in 0.5 bar steps. Although the digital twin and simulation framework accurately model the real process, the system has not yet been validated on an actual industrial setup. Real-world conditions such s sensor delays, actuator nonlinearities, and network latency- may introduce uncertainties that differ from the simulated environment. 

Compared with conventional PID and model-based controllers, the proposed fuzzy expert system offers greater adaptability to nonlinear and uncertain process dynamics. Unlike traditional control methods that require precise mathematical models, the fuzzy logic approach utilizes expert knowledge to maintain process stability even when system parameters fluctuate. This makes the system particularly suitable for real-world industrial environments, where operating conditions and sensor behavior often vary unpredictably.

%future works and outcomes
For future work, it is planned to compare the proposed control approach with existing solutions in systems control and to extend the web application to establish a continuous connection between the digital twin and the user. Currently, direct communication with the digital twin is implemented via MATLAB using the OPC protocol. However, these tasks will be addressed in a separate research project.

\textbf{Funding }

No funding was received for this research.

\textbf{Data availability} 
The datasets generated during the current study are available from the corresponding author upon reasonable request.

\textbf{Declarations}

\textbf{Conflict of interest }

The authors have no conflicts of interest to declare.

\bibliography{sn-bibliography}

\end{document}